%% file: main.tex
\documentclass[sn-nature]{sn-jnl}% Style for submissions to Nature Portfolio journals
\usepackage{graphicx}%
\usepackage{multirow}%
\usepackage{amsmath,amssymb,amsfonts}%
\usepackage{amsthm}%
\usepackage{mathrsfs}%
\usepackage[title]{appendix}%
\usepackage{xcolor}%
\usepackage{textcomp}%
\usepackage{manyfoot}%
\usepackage{booktabs}%
\usepackage{algorithm}%
\usepackage{algorithmicx}%
\usepackage{algpseudocode}%
\usepackage{listings}%
\usepackage{xcolor} 
\newtheorem{definition}{}%

\renewcommand{\thesubsection}{Supplementary Note \arabic{subsection}}

\usepackage{xr}
\usepackage{ulem}
\makeatletter
\newcommand*{\addFileDependency}[1]{% argument=file name and extension
  \typeout{(#1)}
  \@addtofilelist{#1}
  \IfFileExists{#1}{}{\typeout{No file #1.}}
}
\makeatother

\newcommand*{\myexternaldocument}[1]{%
    \externaldocument{#1}%
    \addFileDependency{#1.tex}%
    \addFileDependency{#1.aux}%
}

\myexternaldocument{appendix}
\begin{document}

\title[Article Title]{A Generative AI Technique for Synthesizing a Digital Twin for U.S. Residential Solar Adoption and Generation}
%Synthetic Residential Solar Identification and Generation Profiles for the United States


\author*[1,2]{\fnm{Aparna} \sur{Kishore}}\email{ak8mj@virginia.edu}

\author[3]{\fnm{Swapna} \sur{Thorve}} \email{st6ua@virginia.edu}

\author*[1,2]{\fnm{Madhav} \sur{Marathe}}\email{marathe@virginia.edu}

\affil[1]{\orgdiv{Department of Computer Science}, \orgname{University of Virginia}, \orgaddress{\city{Charlottesville}, \state{Virginia}, \country{United States}}}

\affil[2]{\orgdiv{Biocomplexity Institute}, \orgname{University of Virginia}, \orgaddress{\city{Charlottesville}, \state{Virginia}, \country{United States}}}

\affil[3]{\orgname{Amazon Robotics}, \orgaddress{\state{Massachusetts}, \country{United States}}}


\abstract{Residential rooftop solar adoption is considered crucial for reducing carbon emissions. The lack of photovoltaic (PV) data at a finer resolution (e.g., household, hourly levels) poses a significant roadblock to informed decision-making. We discuss a novel methodology to generate a highly granular, residential-scale realistic dataset for rooftop solar adoption across the contiguous United States. The data-driven methodology consists of: ($i$) integrated machine learning models to identify PV adopters, ($ii$) methods to augment the data using explainable AI techniques to glean insights about key features and their interactions, and ($iii$) methods to generate household-level hourly solar energy output using an analytical model.
The resulting synthetic datasets are validated using real-world data and can serve as a digital twin for modeling downstream tasks.
Finally, a policy-based case study utilizing the digital twin for Virginia demonstrated increased rooftop solar adoption with the 30\% Federal Solar Investment Tax Credit, especially in Low-to-Moderate-Income communities.
}




\maketitle

\input{introduction}

\input{results}
\input{conclusion}
\input{method}

\bibliography{solar} % common bib file
%\input sn-article.bbl

\input{appendix.tex}

\end{document}

%% file: introduction.tex
\section*{Introduction}
\label{sec:introduction}

The emergence of distributed energy generation sources, such as rooftop solar and wind turbines, is propelling a clean energy wave across the United States. 
Riding on rapidly expanding positive public opinion, augmented by support through government incentives and other funding initiatives, there is a growing consensus for their pivotal role in enhancing energy security and reducing emissions~\cite{clean_white, epa_distributed, meya2021renewable}.
By encouraging the widespread adoption of solar panels, the energy industry can create new market opportunities by not only tapping into a clean and renewable source of power but also reducing the reliance on fossil fuels, which are argued to be one of the major contributors to greenhouse gas emissions. This paradigm shift towards solar energy aligns well with global efforts to mitigate the impacts of climate change and carbon emission reduction by promoting sustainable energy practices~\cite{hu2016impact,ebhota2020fossil}. 

There is overwhelming support in the scientific literature for transitioning to renewable energy sources. Projects such as  DeepSolar~\cite{yu2018deepsolar,wang2022deepsolar++} and Google Project Sunroof~\cite{sunroof} have been at the forefront of constructing comprehensive spatiotemporal datasets using historical satellite and aerial imagery. The DeepSolar project, for instance, mapped photovoltaic (PV) installations from 2006 to 2017 across 420 U.S. counties. Recently, Wussow et al.~\cite{wussow2024exploring} further enhanced this database to include installations for all 50 states and the District of Columbia, in the year 2022. 
 
Building upon these foundational works, researchers have also analyzed rooftop solar adoption's current status and future potential at different spatial and temporal resolutions~\cite{LEMAY2023113571,lee2019deeproof,zhang2016data}. In addition to the technical and geographic analyses, research in this domain also assesses the importance of rooftop solar adoption~\cite{shi2024climate} and analyzes the social, demographic, and economic disparities along with their influence on rooftop and community solar adoption, examining various temporal resolutions to understand how adoption rates evolve over time~\cite{barbose2021residential,reames2020distributional,sunter2019disparities,o2024evaluating}. 
While these studies make a compelling case for solar panel adoption, the lack of PV data at a granular resolution (e.g., household and hourly levels) poses a significant roadblock for informed decision-making in optimal investment and incentive designs, policy decisions, grid stability management and accurate forecast modeling. We address this key data and knowledge gap through this work. Our aim is to explore the geographical and temporal dynamics of solar adoption and PV generation using a synthetic population of the U.S.

In our study, we develop a  novel generative AI methodology to assign PVs to households in the U.S., and then create household PV energy profiles. The methodology consists of two key steps. The first step involves using integrated machine learning (ML) models for predicting household solar adoption. We employ an ensemble of ML models for the pre-processing step of square footage classification, followed by a calibrated decision threshold with a custom loss function XGBoost model for solar adopter identification. The model's parameters are chosen through Bayesian optimization to refine the precision of solar adopter identification. Our dataset is validated against real-world data for each state to mirror actual solar adoption patterns~\cite{barbose2021residential}. Furthermore, we utilize explainable artificial intelligence (XAI) techniques using SHapley Additive exPlanations (SHAP)~\cite{lundberg2017unified} to unravel patterns from state-level ML models, offering a transparent and comprehensible analysis of the factors influencing solar adoption~\cite{machlev2022explainable,sarp2021interpretable,kuzlu2020gaining}.

\begin{figure}[!h]
    \centering
    \includegraphics[width=1.0\textwidth]{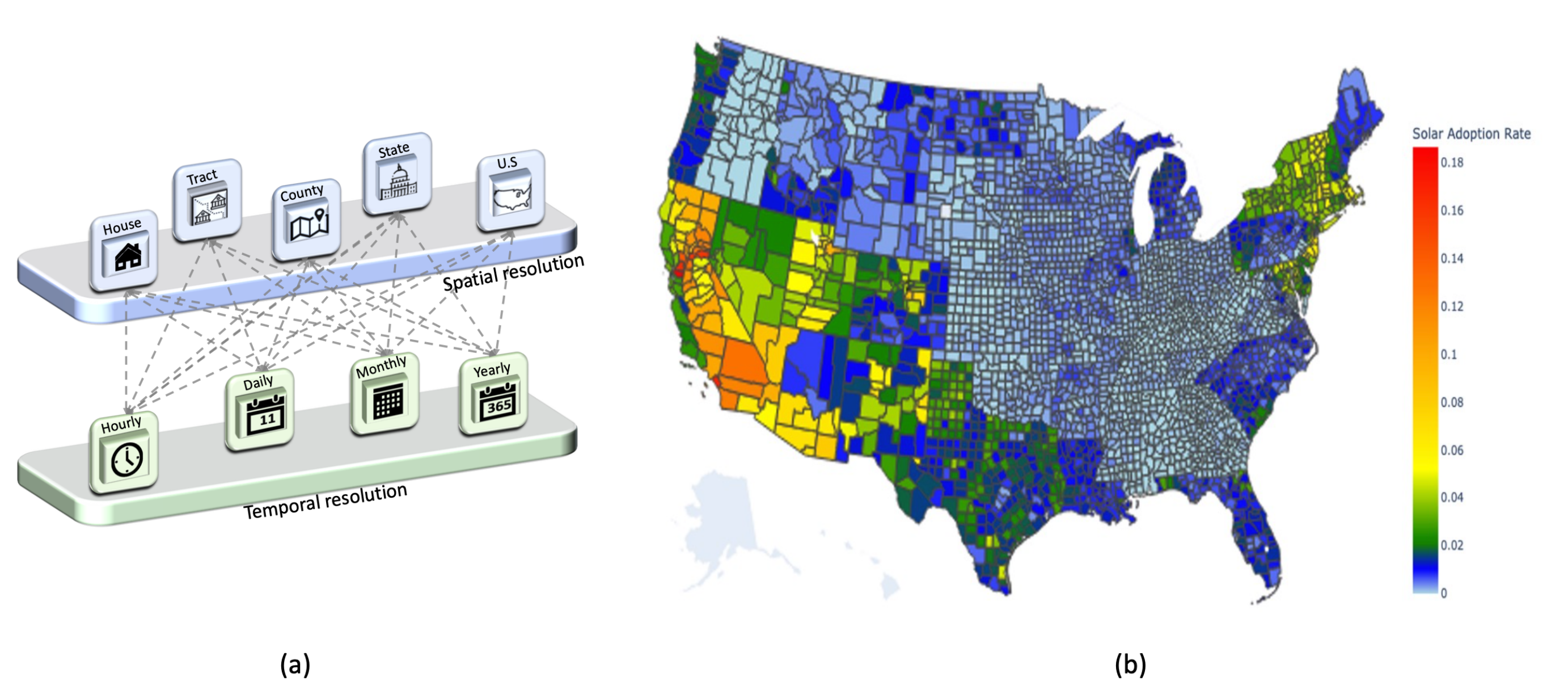}
    \caption{\textbf{U.S. and rooftop solar adoption.} (a) Different combinations of spatial (household, census tract, county, state and U.S.) and temporal (hourly, daily, monthly, yearly) resolutions possible using the solar energy generation model developed in this work. (b) U.S. county-level solar adoption rate choropleth map in the synthetic population. Each county is shaded with the color intensity reflecting the adoption rate. The total solar adoption in each county has been normalized with respect to the number of households in that county. This normalization allows for a more accurate representation of solar adoption rates, as it accounts for variations in county population sizes. The varying intensities of color represent geographical disparities in solar energy uptake across the country. California stands out from other states, exhibiting a significantly higher rate of solar adoption. The map also provides insights into regional trends where the states in the West lead in solar adoption, followed by the Northeast. In addition, the map indicates that the South lags behind the West regarding solar adoption.}
    \label{fig:us_county_adoption}
\end{figure}

In the second step, we employ a bottom-up approach to generate hourly solar production using an analytical model, where we have focused on the detailed aspect of hourly rooftop solar energy production for households across the U.S. In addition, we quantify uncertainties associated with each household for every hour. This method enables us to produce data with high spatial and temporal resolution, facilitating aggregation to broader scales, from households to census tracts, counties, states, and the entire country, or to any temporal resolution, including hourly, daily, weekly, monthly, or specific years (Figure~\ref{fig:us_county_adoption}a). 
%This is also demonstrated in our results. 
The dataset is validated against existing real-world datasets from Lawrence Berkeley Lab (LBNL) residential solar project~\cite{barbose2021residential}, DeepSolar project~\cite{wang2022deepsolar++,wussow2024exploring} and Pecan Street~\cite{pecan_street}. 
Finally, using our framework and the datasets, we present a case study on the effect of different policies in the penetration of rooftop solar adoption in the Commonwealth of Virginia (VA). Our analysis has three key components: individual characteristics, peer effects, and spatial factors, thereby offering a holistic view of solar adoption penetration. We also analyze the distribution of solar adoption across Low-to-Moderate-Income (LMI) and non-LMI communities in both rural and urban settings. This analysis provides valuable insights into the relationships between rooftop solar adoption and local socioeconomic and demographic characteristics.

Our model development primarily relies on open-source datasets and national surveys. 
%Beyond application to a synthetic population, 
Our framework can generate solar production information solely through the use of the Residential Energy Consumption Survey (RECS)~\cite{RECS} and the National Renewable Energy Laboratory (NREL)'s solar irradiance dataset~\cite{nsrdb}. This approach allows for broad applicability and adaptability, ensuring that the model can be effectively utilized in various contexts without needing proprietary data sources. Furthermore, we publish our large-scale synthetic household-level datasets and make them publicly available for research and development purposes.

%% file: results.tex
\section*{Results}
\label{sec:results}

\subsection*{Solar adoption and PV generation: Geographical and temporal dynamics}
\label{subsec:spatio-temporal}

Spatial and temporal insights on rooftop solar adoption and PV generation are critical for sustainable urban development. 

\begin{figure}[!h]
    \centering
    \includegraphics[width=0.99\textwidth]{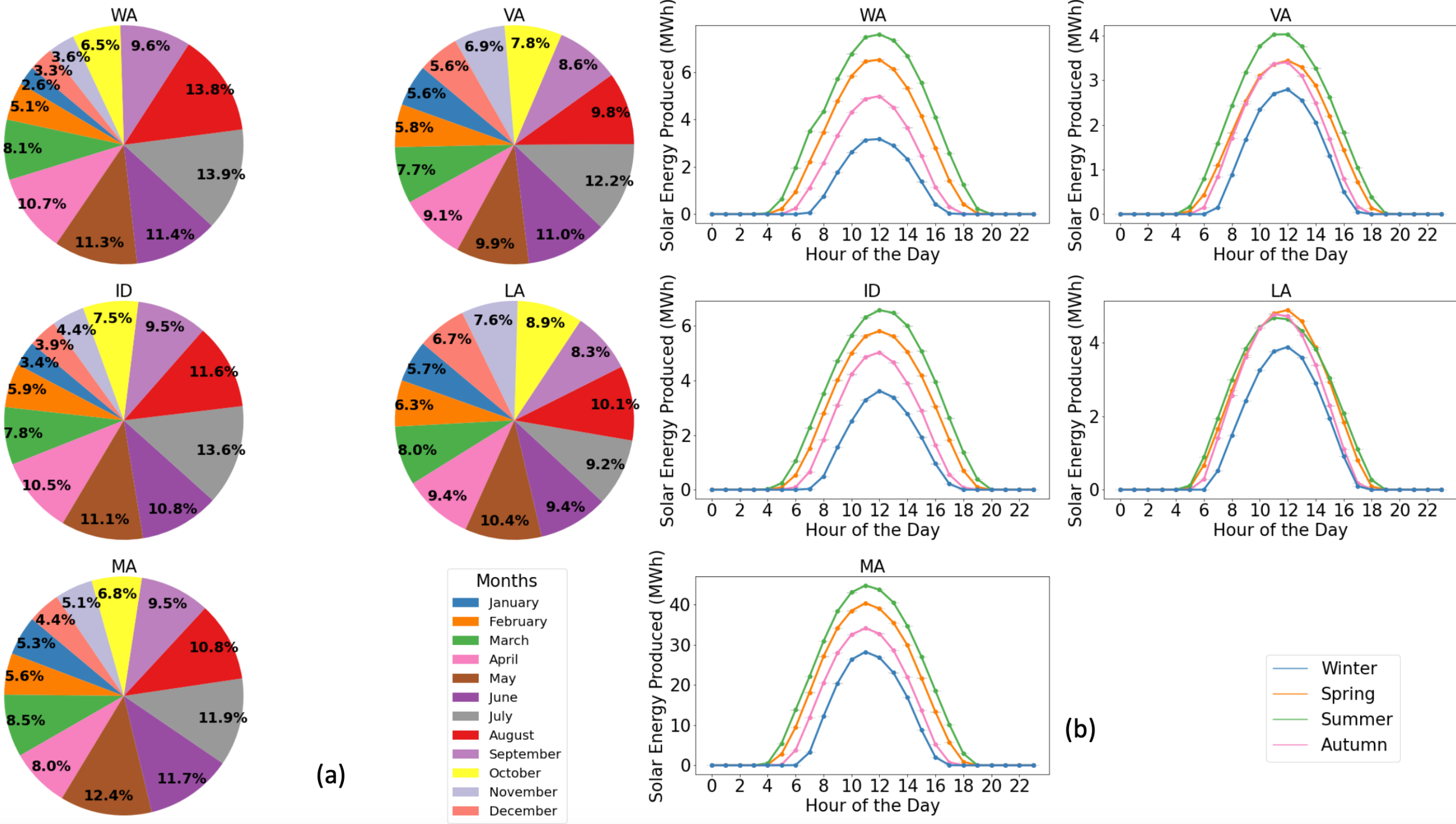}
    \caption{\textbf{Spatial and temporal analysis of solar energy production for WA, VA, ID, LA, and MA in the synthetic population. } (a) Monthly solar energy production: Each pie chart shows the distribution of solar energy generated by all the households across five selected states for each month. It is divided into twelve segments, each corresponding to a month of the year. (b) Hourly aggregate solar energy production by season: The line graph presents the aggregate solar energy produced in each hour by rooftop solar panels for each season. Each data point on the graph represents the total energy produced during a specific hour, aggregated over an entire season. The x-axis indicates the hour of the day with respect to their specific time zones, while the y-axis denotes the hourly-seasonal aggregate solar energy produced, measured in megawatt-hours (MWh). The visualization offers insights into the geographic and temporal fluctuations in residential solar energy generation, reflecting the impact of regional climatic conditions and other environmental factors.}
    \label{fig:spatial_temporal_visualization}
\end{figure}
As an initial step, we identify solar adopters in each state of the U.S. 
%using the 
(methodology is outlined in Section~\ref{subsec:identification} and the pre-processing steps along with the performance metrics are given in Supplementary Note 2 and Supplementary Note 3). We then perform county-level solar adoption visualization that offers insights into the geographic distribution of solar energy adoption across the U.S., highlighting areas with both high penetration and untapped potential. Results of the county-level adoption rates are shown in Figure~\ref{fig:us_county_adoption}b. Solar adoption in each county is normalized with respect to the number of households in that county. This provides a comparative view of solar adoption relative to the population in each region. The Western states, particularly counties in California (CA), exhibit significantly higher adoption rates, followed by the Northeast. The map also reveals that the South trails behind the West regarding solar adoption. This observed pattern aligns with the general solar adoption trends in the U.S.~\cite{county_solar}.

Next, we summarize the solar energy production pattern across different geographic and temporal resolutions in Figure~\ref{fig:spatial_temporal_visualization}. We selected Washington (WA), Virginia (VA), Idaho (ID), Louisiana (LA), and Massachusetts (MA) to act as representative states for the U.S. as they showcase diverse geographic regions, climates, and population densities, providing a comprehensive view of spatial-temporal dynamics across the country. Figure~\ref{fig:spatial_temporal_visualization}(a) provides insights into month-to-month fluctuations in residential solar energy generation, influenced by geographic location and seasonal climatic variations. For instance, WA and ID demonstrate significant variations in their solar energy contributions during winter, likely influenced by their colder climate zones. July has the highest solar energy production for VA, ID, and WA, although LA and MA peak in May. Despite varied winter lows, the minor differences in solar output from spring through fall suggest a reliable level of solar energy generation during these periods, which can significantly reduce energy bills and offset carbon emissions. Figure~\ref{fig:spatial_temporal_visualization}(b) compares hourly solar energy production by season, showing distinct seasonal curves for WA, MA, and ID. VA's spring and autumn curves overlap until noon, illustrating longer solar production in summer. LA's curves overlap, with winter morning production being the lowest. MA stands out with the highest solar energy production, nearly ten times that of other states, due to its large number of solar adopters.
We provide more insights at different resolutions on the spatial and temporal dynamics in Supplementary Note 4. 

\subsection*{Explainability of the models}
\label{subsec:explainability}

\begin{figure}[!h]
    \centering
    \includegraphics[width=0.95\textwidth]{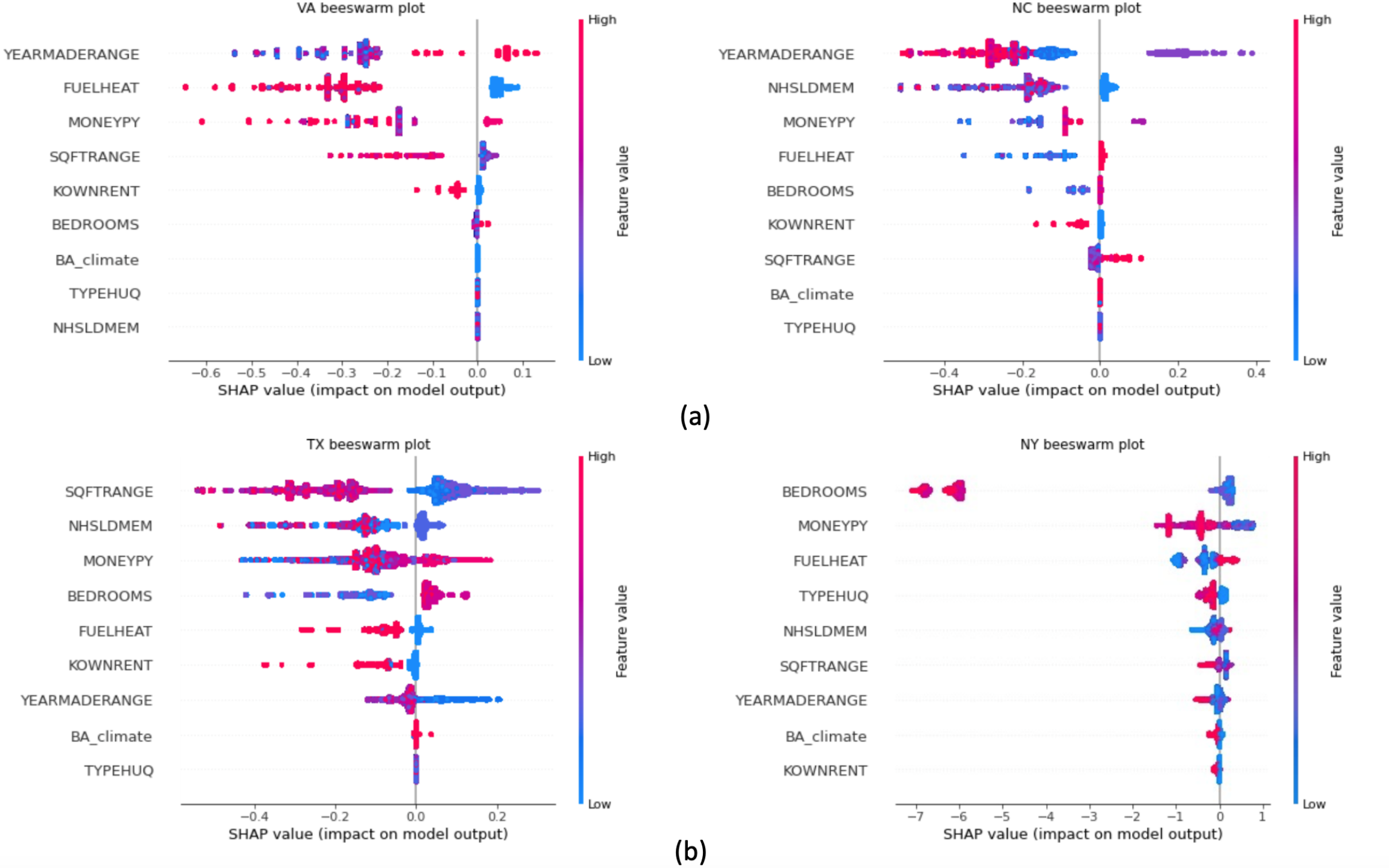}
    \caption{\textbf{Comparative Analysis of State-Level Models at global level using beeswarm plots.} The plots illustrate the SHAP values for various features, arranged on the y-axis according to their importance. The x-axis displays the SHAP values, with color intensity varying from blue to magenta to represent feature values from low to high. Points cluster where data concentration is highest. (a) VA and NC: This figure presents a side-by-side comparison of bee swarm plots for VA and NC. (b) TX and NY: This figure provides a side-by-side comparison of beeswarm plots for TX and NY.}
    \label{fig:explainability}
\end{figure}

Understanding the factors that govern our state-level solar production models' predictions is important for its validation. The model's explainability becomes vital, especially when considering the practical implications of deploying such models in policy-making and energy management. SHAP is a popular explainability technique used in machine learning to interpret the predictions of ML models. It assigns each feature an importance value (SHAP value) based on its contribution to the prediction.

In Figure~\ref{fig:explainability}, we purposefully selected the state-level solar adoption ML models for VA-North Carolina (NC) and Texas (TX)- New York (NY) state-level models to highlight their geographic, political, and climatic similarities/contrasts. The variables used and their descriptions are mentioned in Section~\ref{subsec:classification}.
 In the VA-NC model [Figure~\ref{fig:explainability}(a)], houses constructed recently (YEARMADERANGE) have high SHAP values for VA, suggesting their high contribution to solar adoption. Conversely, NC shows mid-range YEARMADERANGE with higher SHAP values. In this model, recent constructions negatively impact predictions. VA finds medium-sized properties and households with higher income (MONEYPY) positively influencing SHAP values, linking them to solar production. In NC, larger homes and middle incomes positively affect SHAP values, differing from VA's pattern.
 
In the TX-NY model [Figure~\ref{fig:explainability}(b)], a notable contrast is observed in the beeswarm plots for MONEYPY, where the direction and magnitude of this feature contribution appear to reverse between the two states. Property size significantly influences solar adoption predictions in the TX model, while in the NY model, household income and the number of bedrooms are identified as key determinants. Additional analysis for VA-NC and TX-NY models are presented in Supplementary Note 5. Thus, the contrasting SHAP value trends in the state-level models uncover the importance of property-specific details (e.g., square footage range and construction period), demographic characteristics (e.g., number of household members), and socio-economic factors (e.g., income level) across different geographic landscapes in solar adoption. 

\subsection*{Validation of synthetic solar data} 
\label{subsec:validation}

\begin{figure}[!h]
    \centering
    \includegraphics[width=1\textwidth]{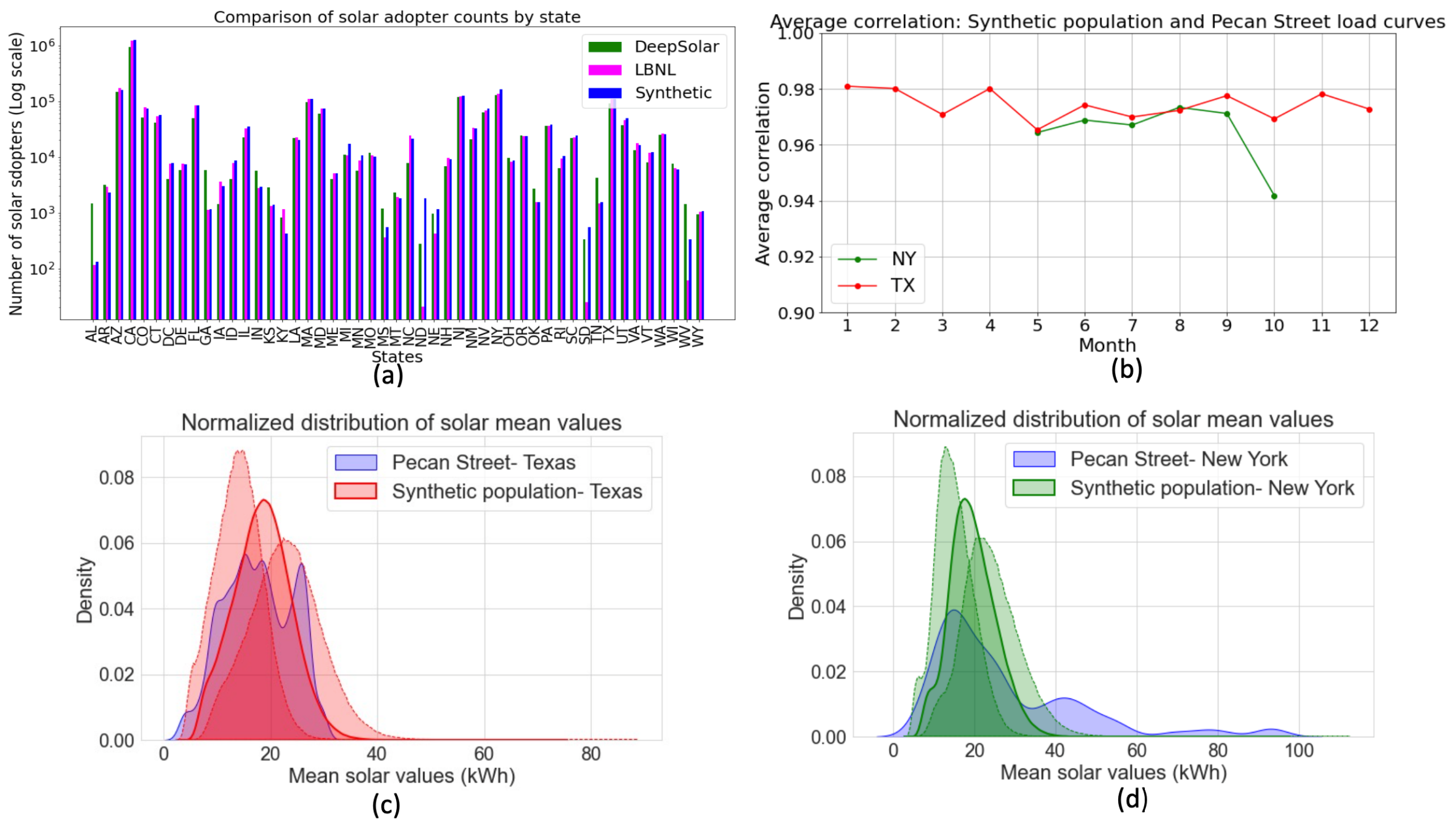}
    \caption{\textbf{Validation of solar adoption and PV generation synthetic datasets.}(a) Comparison of synthetic solar adopters with the DeepSolar and LBNL solar dataset across U.S. states. The x-axis represents the contiguous states in the U.S., while the y-axis denotes the number of solar adopters in log scale. (b) Average correlation between hourly load curves of synthetic households and Pecan Street households. The x-axis represents the month, and the y-axis shows the average correlation calculated using mean Pearson correlation coefficients. The data consistently exhibits a high positive correlation across all months for both TX and NY. (c) Comparison of the daily average solar generation distribution between Pecan Street dataset for Austin, TX and synthetic solar generated dataset for TX.  Pecan Street data is depicted by the solid blue curve. The solid red curve illustrates the mean of the synthetic data, and the red dotted curves indicate the standard deviation of the synthetic dataset. (d) Comparison of the distribution between Pecan street dataset for NY and synthetic solar generated dataset. Pecan Street data is depicted by the solid blue curve. The solid green curve illustrates the mean of the synthetic data, and the green dotted curves indicate the standard deviation of the synthetic dataset.} 
    \label{fig:validation}
\end{figure}

The primary goal is to compare synthetic solar adopter state-wise counts with real-world datasets developed by multiple researchers to assess the accuracy of the distribution against actual solar adoption. Next,  we  compare the load shape curves and PV energy profile distribution between real and synthetic data to validate the temporal and spatial patterns of solar energy generation, ensuring the synthetic models accurately reflect real-world dynamics. We use the LBNL solar project~\cite{barbose2021residential} dataset for residential solar project and the DeepSolar dataset~\cite{wussow2024exploring} with residential state-wise solar adopter count to validate our 
%dataset's 
overall count of adopters. We corroborate our solar generation profiles through the samples of recorded household data from the Pecan Street, updated through 2020, for TX and NY~\cite{pecan_street}. We utilize each household's 15-minute average solar generation data to validate the hourly solar generation load shapes and compare the state-wide distribution.

In Figure~\ref{fig:validation}a, we compared the synthetic solar adopter counts with two residential solar adopter datasets. The results show that the synthetic solar adopter counts are analogous to the two real-world residential solar adopter datasets. More detailed insights on the solar adopter validation are presented in Supplementary Note 6.
Next, we compared daily load patterns between the Pecan Street and our synthetic dataset for TX and NY, selecting an equivalent number of households through random sampling from the synthetic dataset. 
The average monthly Pearson correlation coefficients calculated between pairs of randomly selected synthetic households and Pecan Street households for every hour are shown in Figure~\ref{fig:validation}(b) and they reveal a consistently high positive correlation for both states across all months. 
We compare solar generation distributions from Pecan Street data and synthetic data in kWh in Figures~\ref{fig:validation}(c) and \ref{fig:validation}(d). In Figure~\ref{fig:validation}(c), the Pecan Street dataset is represented as a solid blue curve. The mean and standard deviation of the synthetic data are represented by solid and dotted red curves, respectively, enveloping the blue curve. 
Jensen Shannon Divergence (JSD) values for TX's distributions are 0.18 (histogram) and 0.42 (KDE). 
In Figure~\ref{fig:validation}(d), the synthetic data is shown in green, comparing NY's distributions. The JSD values are 0.39 (histogram) and 0.12 (KDE). 

\emph{Summary:}~The synthetic dataset robustly represents real-world solar generation patterns and solar adopter behaviors across states. JSD values indicate a satisfactory alignment between the synthetic and real-world distributions despite some discrepancies in the density and coverage of outliers. Moreover, the consistently high positive Pearson correlation coefficients for daily load patterns across all months affirm the synthetic data's accuracy.

\subsection*{Case study: Policy impacts on rooftop solar adoption} 
\label{subsec:case_study}

\begin{figure}[!h]
    \centering
    \includegraphics[width=1\textwidth]{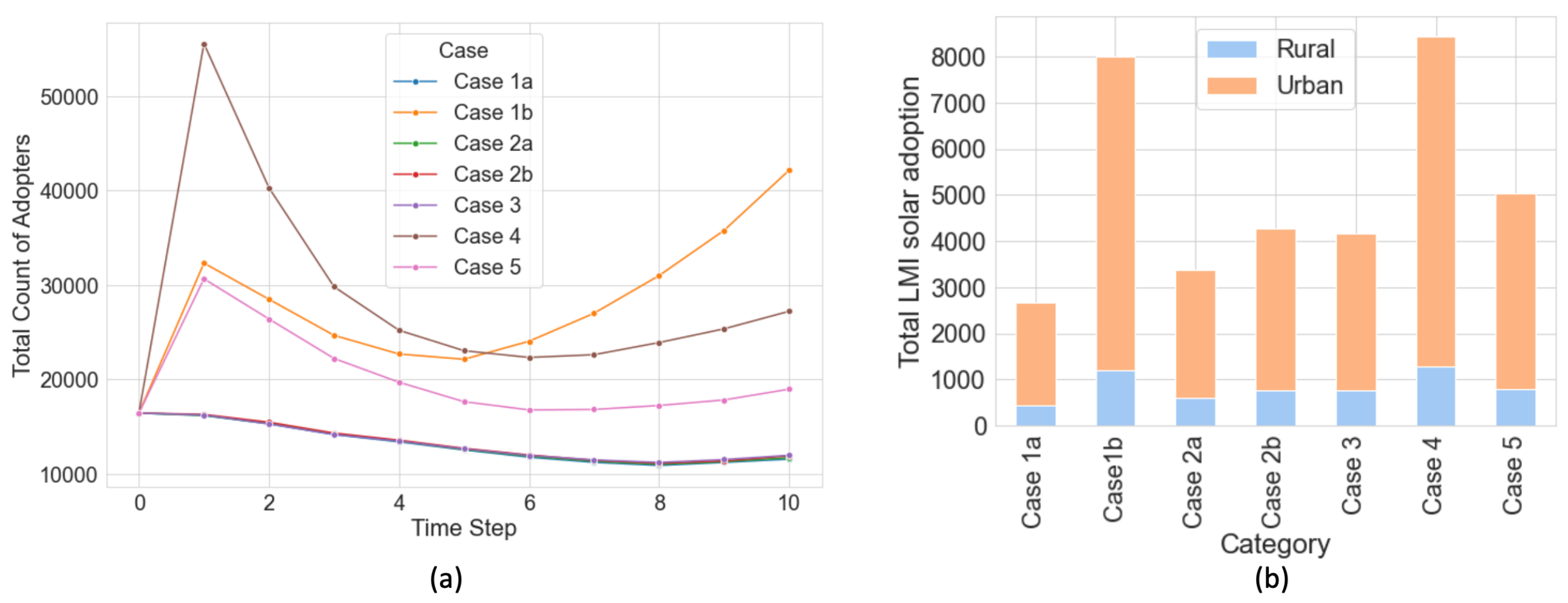}
    \caption{\textbf{Policy impacts on rooftop solar adoption in VA synthetic population.}(a) Comparison of different cases in total solar adoption across VA. The line plot shows adopter counts for seven policies (Cases 1a, 1b, 2a, 2b, 3, 4, and 5) over 10 time steps, with each line depicting a different policy's impact on adoption rates. The x-axis indicates time steps, and the y-axis represents adopter counts. The plot reveals how each policy influences adoption patterns, with Cases 4, 1b, and 5 showing distinct trajectories compared to the similar patterns of Cases 1a, 2a, 2b, and 3. (b) Bar chart of LMI solar adoption in VA's rural and urban areas under various cases. The rural and urban  LMI population is around 31.7\% and 68.3\%, respectively, of the total LMI population. The x-axis lists the cases, and the y-axis shows total LMI solar adoption, with blue for rural and orange for urban areas. Equal opportunity policies (Cases 4 and 1b) show similar rural adoptions, with targeted policies (Cases 2b and 5) following. Case 4 leads in urban adoption, highlighting the 30\% tax credit's effectiveness in enhancing urban LMI penetration.}
    \label{fig:case_study}
\end{figure}

As an illustration of the use of the generated digital twin, we carried out a case study motivated by recent policy considerations. The objective of case study is to use the digital twin to understand the dynamics of solar adoption and its penetration under various case scenarios. 
The scenarios primarily vary in the likelihood of solar adoption for different communities, illustrating the policy interventions. The case scenarios are described in Section~\ref{subsec:approach_casestudy}. 
One of the top challenges in rooftop solar adoption is overcoming the economic barriers. Federal policies, including tax rebates and incentives, play a crucial role in providing financial assistance and thus accelerating clean energy adoption. Additionally, policies focused on increasing solar adoption and penetration in LMI communities help to address social disparities and support the government's objective of the Justice40 initiative~\cite{justice40}. However, evaluating the benefits of these policies and the resulting solar adoption penetration remains challenging. Factors such as peer influence, where social networks influence household decisions, and the social-demographic attributes of the households introduce uncertainty in the penetration of solar energy. Agent-based modeling using digital twins is one of the methods to simulate and better understand this behavior.

Our analysis examines how these different policies, under the influence of peer effects, and microelements such as socio-demographic attributes of households contribute towards solar adoption within the state of VA using a framework for contagion simulation modeling~\cite{priest2021csonnet}, which capture individual components, community components for spatial effects, and neighbor components for the influence of immediate neighbors. We examine the impact of these policies on different segments of the population - ($i$) LMI~\cite{lmi_income} and non-LMI communities, and ($ii$) rural and urban populations. Model design and experimental setup are explained in Section~\ref{subsec:case_study}.

Figure~\ref{fig:case_study}a summarizes our policy study results and shows overall solar adoption trends: Cases 4, 1b, and 5 have unique patterns, while Cases 1a, 2a, 2b, and 3 follow similar paths. The 30\% tax credit policy leads to the highest adoption in all communities. Figure~\ref{fig:case_study}b reveals that equal opportunity policies are effective in urban and rural LMI areas, influenced by community and network structures.  This suggests the significant role of community effects and peer influence in adoption, with non-LMI adopters notably impacting overall penetration. Our case study suggest that policy interventions, like tax credits to all individuals, are essential for higher adoption. 

However, even the most successful policies see less than 1\% LMI rooftop adoption. This outcome is attributed to the chosen network's workplace-based nature. Given the higher unemployment rates within the LMI community~\cite{bls}, this network model results in fewer connections among LMI households, limiting the spread of adoption through peer influence. This study, using the synthetic population of Virginia, demonstrates how digital twins can be utilized to analyze solar adoption and penetration. Similar studies can be conducted in other states or over different networks to gain deeper insights on solar penetration.  

%% file: conclusion.tex
\section*{Discussion}
\label{sec:conclusion}

Our work has bridged a key research gap
%significantly contributed 
in the understanding and prediction of rooftop residential solar adoption work. We identified solar adopters using a calibrated XGBoost model, which
%for precise outcomes. 
was validated against real-world adoption data, ensuring its reliability in reflecting adoption patterns across different states.
% 88\% of the U.S. contiguous states have the state-level adopter difference between real-world and synthetic adopters under 25\%. 
Our county-level adoption rate analysis showed that the Western states have significantly higher adoption rates. The Northeast followed this trend next. 
%The South trails behind the West regarding solar adoption.
This observed pattern aligned well with the general solar adoption trends in the U.S.~\cite{county_solar}.

Next, we developed a comprehensive model to simulate the hourly solar energy production profiles of residential rooftop PV systems across the U.S. The data generated from the model provided a detailed account of the energy outputs throughout the year and offered insights into the spatial and temporal variations. We extensively validated the generated datasets against real-world datasets based on their aggregate PV energy distribution and daily load shape patterns. Furthermore, our framework has been developed to run for different temporal resolutions (e.g., day, week, and month). The synthetic solar adoption and PV energy generation dataset served as a digital twin for residential rooftop solar adoption in the U.S., playing a vital role in devising interventions and policy strategies to improve rooftop PV adoption. 
%We use parallel cluster computing using Message Passing Interface (MPI) for scalability and efficiency.  

Our study, while comprehensive, also has some limitations. First, the model generated the suitable area for solar adoption based on the roof area of the house as described in the literature. Hence, the model did not account for extreme cases of solar panel installation that could occur in real life. Additionally, our analysis did not consider solar panels installed at different times within the same year. The solar energy generated was calculated based on solar radiation incident on the tilted panels. Other incidences were not calculated, such as the reflected and diffused radiation. 
We also do not account for solar panels present in multiple tilts or azimuth directions simultaneously in a household.
However, one of the ways we tried to address these limitations was by estimating the mean and standard deviation of hourly PV energy profiles. 
Our models assumed that the house owners address shading by trees and the roof's suitability for solar panel installation, which can significantly impact solar energy production.

Through XAI, we have shed light on the feature contributions and interactions within our predictive state-level solar adoption models. We have made our large-scale, fine-grained hourly household-level PV energy profile datasets available for future research. 
A case study on VA rooftop solar adoption illustrated the utility of digital twinning \cite{douglass2024swimming} by examining the impacts of various policies on solar penetration across different communities. This study explored individual, peer, and spatial factors influencing adoption, providing a nuanced understanding of how socioeconomic and demographic factors correlate with solar adoption rates in both non-LMI and LMI communities within urban and rural settings. Thus, our research provided a holistic view of residential rooftop solar panel adoption involving residential solar adopters and their PV energy profile generation at finer resolutions.
%XAI for the state-level model comparison; and a case study that focuses on the individual, peer, and social influence and other socioeconomic aspects on solar adoption.

%% file: method.tex
\section*{Methods}
\label{sec:methods}
\subsection*{Solar adoption and PV profile generation}
\label{subsec:approach_solar}

\begin{figure}[!h]
    \centering
    \includegraphics[width=0.99\textwidth]{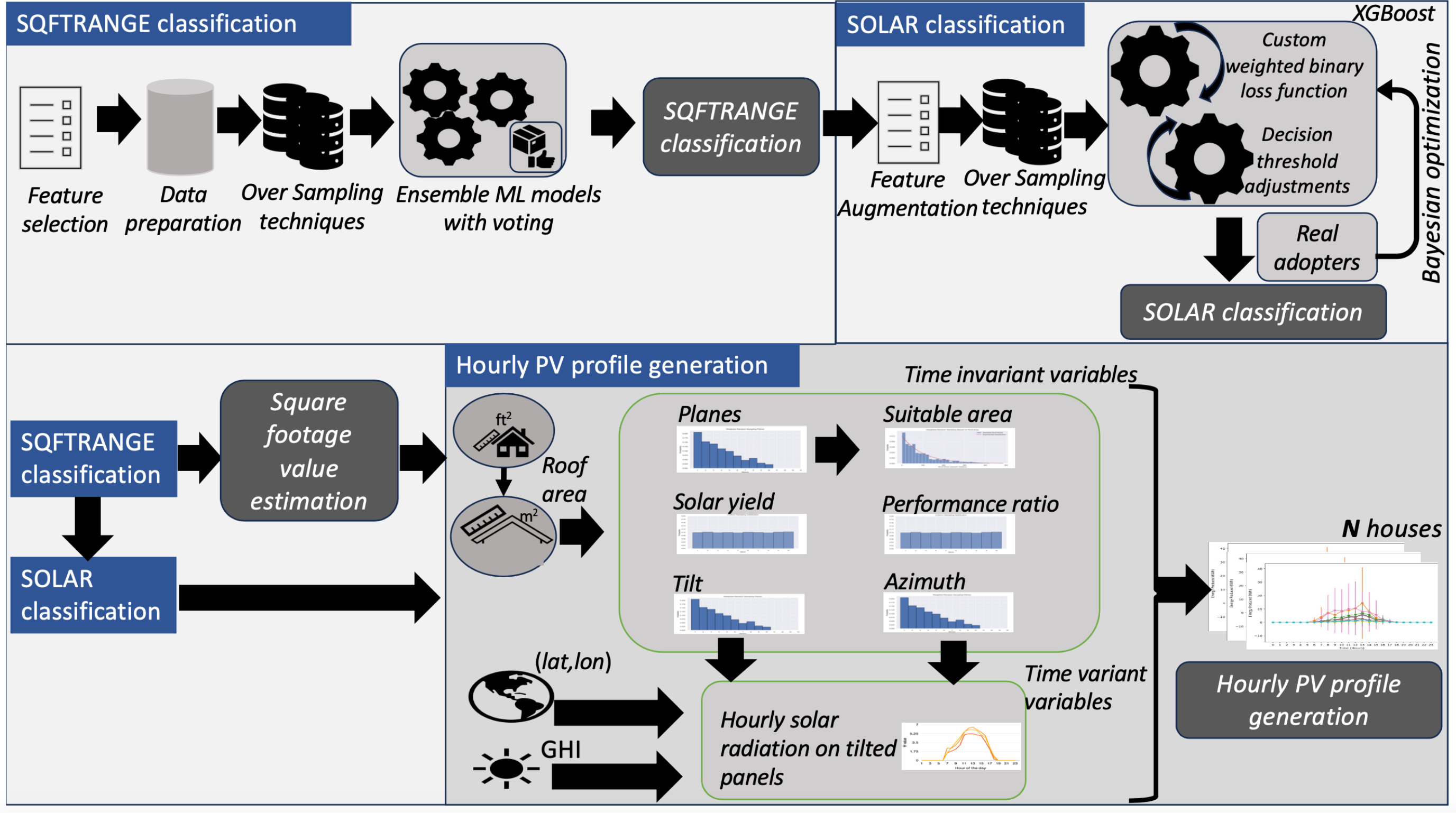}
    \caption{\textbf{Schematic representation of our approach for solar adopter identification and PV energy profile generation.} Step 1: SQFTRANGE classification - The initial steps for SQFTRANGE classification are feature selection and data preparation. Next, we implement an ensemble of classifiers - including a random forest, SVM, and gradient boosting followed by a dual voting mechanism involving soft and hard voting. SQFTRANGE is augmented with the other features for SOLAR adopter identification. Step 2: SOLAR classification - The initial steps resemble SQFTRANGE prediction. Model learning involves an XGBoost classifier with a custom weighted binary loss function and a calibrated decision threshold. A feedback loop from the ground truth solar adopters helps to calibrate the model using Bayesian optimization. SQFTRANGE prediction of each household is used to estimate the square footage of the house. Step 3: Hourly PV profile generation - The next series of steps details the process from assessing roof area to creating the final ensemble of PV energy profiles. These include generating random samples for time-invariant variables, determining suitable areas via weighted and exponential sampling, and calculating solar radiation and energy profiles based on geographical data and irradiance. }
    \label{fig:methods}
\end{figure}

We aim to generate hourly PV energy profiles for households that have adopted solar energy. All datasets, the variables/abbreviations along with their explanations used in this research are mentioned in Supplementary Note 1. We employ the synthetic population dataset~\cite{adiga15US} for this purpose. However, this dataset lacks detailed information on household square footage and solar adoption status. To overcome this limitation, we integrate data from the RECS to enrich the synthetic population with the necessary details on square footage and solar adoption. After identifying the solar adopters, we proceed with the energy profile generation for the solar-adopted households. Our approach for solar adoption and PV energy profile generation is presented in Figure~\ref{fig:methods}. 
Our methodology unfolds in three distinct steps to systematically enhance the dataset and produce reliable PV energy profiles: ($i$) classifying synthetic households into predetermined square footage range categories (Synthetic household square footage range classification), ($ii$) identifying the households within the synthetic population that have solar panels (Identification of solar adopters), and ($iii$) producing hourly PV energy profiles for homes recognized as solar energy adopters (Generation of Hourly PV Energy Profiles). 

\subsubsection*{Synthetic household square footage range classification}
\label{subsec:classification}
The square footage range (SQFTRANGE) classification is treated as a multi-class classification task, categorizing households based on their square footage ranges. 
First, we identify common independent variables (features) associated with the demographics and household properties across both the RECS survey and the synthetic households. We employ Recursive Feature Elimination with Random Forest classifier to determine the minimal set of essential features for the square footage class prediction. The final feature set includes the number of household members (NHSLDMEM), number of bedrooms (BEDROOMS), type of housing unit (TYPEHUQ), main space heating fuel (FUELHEAT), ownership status of the unit (KOWNRENT), the year range in which the housing unit was built (YEARMADERANGE), annual household income range (MONEYPY), and the building's climate zone (BA\_climate).  
These features undergo pre-processing to maintain consistent meaning for the categorical values between the two datasets (RECS and synthetic population). Next, we address class imbalance issues in the SQFTRANGE classification. We utilize an oversampling technique, the Synthetic Minority Over-sampling Technique for Nominal (SMOTEN)~\cite{chawla2002smote}. To ensure inter-feature correlations remain similar before and after SMOTEN, we compute the correlation matrix using Cramer's V coefficient~\cite{mchugh2013chi}. 

Next, we train various machine learning models tailored to different census divisions across the U.S., aiming to capture the regional similarities and differences for SQFTRANGE prediction. We employ an ensemble of machine learning models for the task of predicting SQFTRANGE classifications. Our ensemble is comprised of: ($i$) a random forest classifier, ($ii$) a support vector machine (SVM) classifier, and ($iii$) a gradient boosting classifier. These models have undergone hyperparameter optimization through a grid search. 
For making predictions, we adopt a voting mechanism that takes into account the outputs from all three classifiers. Specifically, we employ a plurality/hard voting strategy~\cite{leon2017evaluating} when at least two classifiers agree on the predicted class, determining the final class based on this majority consensus. Conversely, when each classifier outputs a different class, we resort to a soft voting approach~\cite{voting}. This method involves aggregating the predicted class probabilities from all classifiers, and the class with the highest total probability is selected as the ensemble's output. This combination of voting strategies enhances the robustness and generalizability of our model, mitigating the biases and overfitting tendencies that might affect the individual classifiers.

Finally, we apply the ensemble trained model tailored for each census division to predict the SQFTRANGE for synthetic households in the corresponding U.S. states within each division. Upon obtaining the square footage predictions, we integrate them with the existing attributes of each household. Following this augmentation, we 
proceed to the next task of identifying solar adopters in the synthetic population.

\subsubsection*{Identification of solar adopters} 
\label{subsec:identification}
The solar adopter identification modeling is performed on a state-wise basis to incorporate state-specific characteristics into the model. We employ over-sampling techniques like random oversampling or SMOTEN to address the class imbalances. Next, we perform model learning using an XGBoost classifier. The rationale for choosing the XGBoost classifier is twofold: first, tree-based classifiers are well-suited for tabular data~\cite{shwartz2022tabular}, and second, XGBoost permits the integration of a custom log loss function and the calibration of the decision threshold, aligning closely with our problem setting. We undertake this step to satisfy our specific objective of minimizing the discrepancy between the actual state-wise solar adopter counts in the U.S. and the predicted counts in our synthetic model. The real state-wise solar adopter counts for the U.S are derived from the data published by LBNL~\cite{barbose2021residential}. Here, the data specifies the percentage and count of state-wise samples, enabling us to approximate the real number of adopters. 

In the weighted custom log loss function, we introduce a penalty term, $\beta$ to allow us to control the penalty given for false positives/false negatives while training. The weighted log loss function is expressed as:
\begin{equation}
\text{Weighted Log Loss} = - \frac{1}{T} \sum_{i=1}^{T} \left[ y_i \log(\hat{y}_i) + \beta \cdot (1 - y_i) \log(1 - \hat{y}_i) \right]
\end{equation}

where $T$ is the total number of samples, $y_i$ is the true label of the $i^{th}$ sample, $\hat{y}_i$ is the predicted probability of the $i^{th}$ sample to belong to positive class and $\beta$ adjusts the weight given to the negative class in the loss function~\cite{chen2016xgboost}. 

We also adjust the decision threshold, $\tau$ while training the machine learning model~\cite{hu2021decision}, such that the binary decision label $D_i$ is predicted as  
\begin{equation}
D_i = 
\begin{cases} 
1 & \text{if } \hat{y}_i \geq \tau \\
0 & \text{otherwise}
\end{cases}
\end{equation}

These customizations assist with the class imbalance problem and also achieve our objective of minimizing the discrepancy. 
Next, the task is to identify optimal $\beta$ and $\tau$ without brute-force hyperparameter search.
We explore the search space using Bayesian optimization method that uses expected improvement as the acquisition function~\cite{jones1998efficient}. In our approach, the Gaussian process regressor (GPR) serves as a probabilistic model in the Bayesian optimization framework. The key feature of a Gaussian Process model is its ability to provide a predictive mean and variance (uncertainty) for any point in the input space, based on the observations made so far. Expected improvement (EI) uses the mean and variance provided by the GPR to determine which new combination of hyperparameters is most likely to yield an improvement over the current best result.

The EI acquisition function is given as 
\begin{equation}
\text{EI}(x) = (f_{\min} - \mu(x)) \Phi\left(\frac{f_{\min} - \mu(x)}{\sigma(x)}\right) + \sigma(x) \phi\left(\frac{f_{\min} - \mu(x)}{\sigma(x)}\right)
\end{equation}
where $f_{\min}$ is the observed current minimum discrepancy value, $\Phi$ and $\phi$ are the cumulative distribution function (CDF) and probability density function (PDF) of the standard normal distribution, respectively, and $\mu(x)$ and $\sigma(x)$ are the mean and standard deviation obtained at the point $x$ given by the Gaussian process regressor model. The first part in the equation captures the expected improvement due to mean predictions lower than $f_{\min}$. The second part in the equation accounts for the uncertainty of the prediction at $x$, encouraging to explore at the points where the model is uncertain. Thus, the model explores the areas of high uncertainty and exploits the regions where the model predicts low discrepancy. This helps in efficiently identifying the minimum value, to satisfy our objective of minimizing the discrepancy. Finally, the best $\beta$ and $\tau$ selected by the GPR model are used to predict the solar adopters in the synthetic population. 

\subsubsection*{Generation of hourly PV profiles}
\label{subsec:generatin}

\noindent \textbf{\label{subsec:estimation}Square footage estimation}: 
The pre-processing step of square footage estimation is essential before proceeding to the final step of generating hourly PV energy profiles to calculate the suitable area for roof-top PV installation. It involves determining the square footage values for each household in the synthetic population. 

To estimate a house's square footage, we systematically divide square footage ranges into sub-classes, each with a weight based on occurrence frequency in RECS survey data. Taking a household $h_i$ as an example, if the predicted square footage range is between $N_{low} - N_{high}$, the sub-classes could be represented as $(N_{low} - N_{1}), (N_{1} - N_{2}), ..., (N_{k} - N_{high})$. Each sub-class has an associated weight, $W_{N1}, W_{N2}, \cdots$, which is derived from the frequency of its occurrence in the RECS survey data. These weighted sub-classes help narrow down the square footage estimate more precisely.

Next, we proceed with weighted random sampling to select $M$ sub-classes yielding, $S_{N1}, S_{N2}, \cdots S_{NM}$. The likelihood of selecting a particular sub-class is proportional to its weight, ensuring that more common sub-classes are more likely to be chosen. Finally, for each of the selected $M$ sub-classes, we uniformly sample $L$ square footage values. This step introduces variability and accounts for the inherent diversity within each sub-class. Consequently, we end up with a total of $M * L$ square footage values for household $h_i$. 
To provide a more stable and reliable square footage estimate for a household, we take the average $M * L$ values.
Mathematically, the final square footage estimate for household $h_i$, denoted as $\hat{SF}_{h_i}$, is given by:

\begin{equation}
\hat{SF}_{h_i} = \frac{1}{M \times L} \sum_{j=1}^{M} \sum_{k=1}^{L} SF_{jk}
\end{equation}
where $SF_{jk}$ represents the $k^{th}$ square footage value sampled from the $j^{th}$ selected sub-class.

We proceed to develop and create hourly PV energy profiles for households that have adopted solar panels. The time-invariant and time-variant variable categories are explained below. 

\begin{definition}
\label{def:time_invariant}
\textbf{Time-invariant variables}: For each solar adopter household $h_i$, the time-invariant variables include $A_i$, representing a possible set of suitable roof areas for solar panels; $\eta_i$, a set of values for solar panel yield; $PR_i$, a set of values for performance ratio; $\theta_i$, a set of possible tilt angles; and $\omega_i$, a set of possible azimuth directions.
\end{definition}

\begin{definition}
\label{def:time_variant}
\textbf{Time-variant variables}: For each solar adopter household $h_i$, the time-variant variable includes $HT_{i,d,w}$, representing hourly solar radiation values on tilted panels for a given household, for household $i$ for a given day $d$ at a given hour $w$. These values are generated based on the geographical coordinates ($lat_i$,$lon_i$) and the hourly global horizontal irradiance ($GHI_{i,d,w}$) for a given day in a given census tract.
\end{definition}

We calculate residential solar energy profiles for each household using the Equation~\ref{eq:generation}.
\begin{equation}
\label{eq:generation}
E_{i,d,w} = A_i \cdot \eta_i \cdot HT_{i,d,w} \cdot PR_i,
\end{equation}
where $E_{i,d,w}$ is the energy ($kWh$) for household $i$ for a given day $d$ at a given hour $w$. 

We begin by calculating the roof area for each of the households based on the the square footage estimated for the household based on the pre-processing step. We approximate the square footage for the roof as 1.5 times the square footage of the house for each of the households. We estimate the potential roof areas suitable for solar panel installations for individual households. Next, we compute the time-invariant variables. 

First, we generate weighted random samples for the number of planes in the roof~\cite{gagnon2016rooftop}. Next, we proceed to calculate the suitable roof area. The households are categorized based on their roof size, and specific parameters~\cite{gagnon2016rooftop} are assigned according to the number of planes in the roof. Utilizing a uniform distribution, we generate a set of roof areas and apply an exponential probability density function to assign weights to these areas, reflecting their suitability for solar installation~\cite{gagnon2016rooftop}. Next, we employ a weighted sampling technique to select a subset of suitable areas. This approach ensures a tailored identification of possible solar installation areas, accommodating the diverse characteristics and orientations of household roofs. We employ a uniform sampling mechanism to select each sample with equal probability for solar yield and performance ratio. We estimate potential roof tilts and azimuth directions for the household by using weighted random sampling mechanism. The weights are assigned based on the distribution of rooftop planes in each tilt and azimuth category building type, as informed by the existing literature~\cite{gagnon2016rooftop}. We calculate the radiation incident on a tilted surface by using the equation~\ref{eq:radiation}~\cite{duffie2020solar}. We also apply a degradation value based on the selected azimuth direction ($\omega_i$).

\begin{equation}
\label{eq:radiation}
HT_{i,d,w} = \left(\frac{GHI_{i,d,w} \cdot \sin\left((90^\circ - \text{lat}_i + \delta\right) + \theta_i)}{\sin\left(90^\circ - \text{lat}_i + \delta\right)} \right) \cdot D(\omega_i)
\end{equation}
\text{where $\delta$ is the declination angle.}
$D(\omega_i)$ represents the degradation factor as a function of azimuth direction ($\omega_i$).

Finally, we compute the household rooftop PV energy hourly profiles using Equation~\ref{eq:generation}. This equation estimates the energy generated by a household PV system, measured in kilowatt-hours for each hour of the day. To analyze the variability and typical performance of these systems, we calculate the mean and standard deviation of the generated energy values. For each hour, we consider all possible energy output values and compute their average (mean) and measure of variability (standard deviation). This statistical approach provides insights into the expected performance of a household PV system under different settings. We describe the algorithms and running time complexity of solar adoption and energy profile generation in Supplementary Note 7.

\subsection*{State selection for results and validation}
\label{subsec:approach_state}

Our study focuses on different states in the U.S., chosen as representative examples for our result analysis and validation purposes. The details are described in Table~\ref{tab:census}. For SQFTRANGE classification and spatial-temporal dynamics study, we selected five states across the U.S. These selected states bring insights into the spatial distribution and geographical nuances across the U.S. as they belong to different U.S. census divisions and climatic zones defined by the International Energy Conservation Code (IECC)~\cite{iecc}. 

For the explainability of the state-level models, we chose VA-NC and TX-NY models. In the VA-NC model, both states have geographic similarities. However, based on the outcomes of the past four presidential elections, they have different political affinities. In the TX-NY model, both states have a comparable number of solar adopters and high adoption rates. However, they differ significantly in geographic and political affiliation. We selected TX and NY for validating the PV energy generated profiles based on the availability of real-world datasets.

\begin{table}[ht]
\label{tab:census}
\centering
\caption{States, U.S. census divisions, climatic zones and usage.}
\begin{tabular}{|l|l|p{2cm}|l|p{2.6cm}|}
\toprule
\textbf{State} & \textbf{U.S. Division} & \textbf{Climatic zone} & \textbf{Coast} &\textbf{Usage}\\
\midrule
Virginia (VA) & South Atlantic & Mixed-Humid & East & SQFTRANGE classification, spatial-temporal analysis, XAI \\
\hline
Louisiana (LA) & West South Central & Hot-Humid, Mixed-Humid & Gulf &  SQFTRANGE classification, spatial-temporal analysis\\
\hline
Idaho (ID) & Mountain North & Cold & No & SQFTRANGE classification, spatial-temporal analysis\\
\hline
Washington (WA) & Pacific & Cold, Marine & West & SQFTRANGE classification, spatial-temporal analysis\\
\hline
Massachusetts (MA) & New England & Cold & East & SQFTRANGE classification, spatial-temporal analysis\\
\hline
North Carolina (NC) & South Atlantic & Mixed-Humid, Hot-Humid, Cold & East & XAI\\
\hline
Texas (TX) & West South Central & Hot-Humid, Mixed-Humid, Hot-Dry, Mixed-Dry & Gulf & XAI, Validation\\
\hline
New York (NY) & Middle Atlantic & Cold, Mixed-Humid & East & XAI, Validation \\
\bottomrule
\end{tabular}
\end{table}

\subsection*{Explainability of the models}
\label{subsec:approach_explain}

We utilize SHAP (SHapley Additive exPlanations)~\cite{lundberg2017unified} to interpret and compare the predictions of these models. SHAP framework aids in understanding the predictions by comparing the contributions of different features to the outcome. We study the feature contribution at a global (state-level) and local level (for every data point). Additionally, we provided insights into the interactions between different features using SHAP.

\subsection*{Validation of dataset}
\label{subsec:approach_validate}

We compared our synthetic solar adopter data with the state-wise residential solar adopter data from LBNL~\cite{barbose2021residential} and DeepSolar~\cite{wussow2024exploring}. We compared the absolute value as well as calculated the relative percentage differences between the datasets.
Our validation approach for energy profile has two steps: ($i$) compare the aggregated generated energy distributions and ($ii$) compare the daily load curve shape. First, we aggregate hourly generation to daily totals and calculate their average and standard deviations for the synthetic data. This process mirrors the Pecan Street data aggregation for average daily generation. We assess differences using the Jensen-Shannon Divergence (JSD), which scores from 0 (same distributions) to 1 (different distributions), providing a symmetric comparison unlike Kullback-Leibler Divergence (KLD). Mathematically, for two probability distributions \( P \) and \( Q \), the JSD is given by:

\begin{equation}
\text{JSD}(P \parallel Q) = \frac{1}{2} \text{KLD}(P \parallel M) + \frac{1}{2} \text{KLD}(Q \parallel M)
\end{equation}
where \( M = \frac{1}{2}(P + Q) \) and the KLD is calculated as:

\begin{equation}
\text{KLD}(P \parallel Q) = \sum_{x} P(x) \log\left(\frac{P(x)}{Q(x)}\right)
\end{equation}

We used two methods to calculate the JSD and compare probability distributions. The first, a histogram-based approach, compares average solar generation, providing a direct comparison. The second method, Kernel Density Estimation (KDE), considers average values and standard deviations, offering insight into the data's variability. Our synthetic dataset includes standard deviation for refined bandwidth in KDE, unlike the Pecan Street data, which uses a default bandwidth determined by Scott's Rule, $h=n^\frac{-1}{5}$, based on the number of data points. Next, we compared daily load patterns between Pecan Street and our synthetic dataset for TX and NY, selecting an equivalent number of households. First we aggregated hourly data, for each month and for each household. Next, for each month, we computed the Pearson correlation coefficient of the aggregated hourly data between the Pecan Street households and the selected synthetic households. Finally, we computed the mean correlation for each month.

\subsection*{Case study}
\label{subsec:approach_casestudy}
Our objective in this case study is to analyze the evolution of solar adoption using digital twin by considering the influence of policies, peer effects, and microelements such as socio-demographic attributes of households. We analyzed solar adoption penetration and the influence of policies on LMI community in urban and rural populations using seven different cases across five unique setups. We simulate rooftop penetration in households using CSonNet, a framework for contagion simulation modeling~\cite{priest2021csonnet}. 

\noindent \textbf{Model and utility function:} Drawing on Bale et al.'s work~\cite{BALE2013833}, we developed a new heterogeneous model in CSonNet to capture technology diffusion. In the heterogeneous model in CSonNet for solar adoption, a household shifts from a non-adopter (0) to an adopter (1) state if its utility, $u_i$, surpasses the node threshold. The state transition is progressive; once a node shifts to state 1, it does not revert to 0. The utility function comprises of an individual component, a community component for spatial effects, and a neighbor component to capture the influence of immediate neighbors. 
This technology diffusion model captures the personal advantage of adopting specific policies along with the neighbor and community influence in contrast to conventional threshold models~\cite{granovetter:ajs1978} used to analyze the spread of social contagions.  The utility function for household $i$ is expressed as:

\begin{equation}
u_i = w_1 \cdot p_i + w_2 \cdot c_i + w_3 \cdot n_i
\end{equation}

Here, $w_1$, $w_2$, and $w_3$ are the weights for personal benefits, community, and neighbor influences, respectively, with $p_i$ indicating personal benefit value, $c_i$ the community adoption rate, and $n_i$ the neighbor adoption rate.

\begin{table}[ht]
\centering
\caption{Parameter values used in CSonNet simulation framework}
\label{tab:parameters}
\begin{tabular}{|l|p{3.5cm}|p{1cm}|p{4.5cm}|}
\hline
\textbf{Parameters} & \textbf{Description} & \textbf{Range}  & \textbf{Chosen values}\\ \hline
Edge probability & likelihood that any given pair of nodes is connected by an edge & [0.0, 1.0] & 0.1 \\ \hline
Time step & Number of steps for which the simulation is performed & $\ge 1$ & 10 \\ \hline
Iteration & Number of simulation runs & $\ge 1$ & 1 \\ \hline
Node probability & Likelihood of solar adoption & [0.0,1.0] & Case 1a: 0.1 \\  & & & Case 1b: 0.2 \\ 
& & & Case 2a: 0.1 (Non-LMI), 0.2 (LMI) \\ 
& & & Case 2b: 0.1 (Non-LMI), 0.5 (LMI) \\ 
& & & Case 3: 0.3, 0.1, 0.15, 0.2, 0.25, 0.3, 0.35, 0.4, 0.45, 0.5 (LMI)~\cite{o2021impact} \\ 
& & & Case 4: [0.0,0.1]~\cite{solar_rebate} \\ 
& & & Case 5: [0.0,0.1] \\ \hline
Node threshold & Barriers for a household to adopt solar &  [0.0,1.0] & [0.1,0.95] \\ \hline
$w_1$ & Weight for personal benefit & [0.0,1.0] & 0.4 \\ \hline
$w_2$ & Weight for community influence & [0.0,1.0] & 0.3 \\ \hline
$w_3$ & Weight for neighbor influence & [0.0,1.0] & 0.3 \\ \hline
$p_i$ & Personal benefit for household $i$ & [0.0,1.0] & [0.1,0.95] \\ \hline
$c_i$ & Solar adoption rate in the county & [0.0,1.0] & [0.0,1.0] \\ \hline
$n_i$ & Neighborhood solar adoption rate & [0.0,1.0] & [0.0,1.0] \\ \hline
\end{tabular}
\end{table}

\noindent \textbf{Experimental setup:} The input for the contagion model framework includes the network, initial state, the configuration and the transition rule. 
The output is the set of households that are adopters in each time step.

\textit{Network:} We generated a peer-network graph~\cite{thorve2022framework} for the state of VA using NetXPipe~\cite{thorve2023network}. 
This network represents a workplace network within the state, with nodes representing the households and edges indicating connections between nodes that work at the same location. 
For this study, we assume a probability of 0.1 for the existence of an edge between any two nodes. 
Although the network probability depends on factors such as organizational structure and team dynamics, a probability of 0.1 models sparse to moderate interactions between individuals in a workplace helps strike a balance between sparsity and density in the network.
The undirected network comprises a total of 3,094,255 nodes and 138,322,576 edges.

\textit{Initial state:} The initial state is the set of households that are current solar adopters identified using the solar adoption model described in Subsection~\ref{subsec:identification}.

\textit{Configuration:} The configuration consists of setting the model parameters, such as the time step, graph type, and number of iterations. In this setup, we are analyzing the adoption trend for 10 time steps using the undirected, peer-network for 1 iteration. We have selected a time step of 10 in order to find a balance between observing immediate changes and allowing the network to evolve over a significant period. This value ensures that we do not overlook short-term dynamics while also avoiding excessively high values that can fail to capture the evolution of adoption due to technological advancements and other interventions. The number of iterations is chosen as 1,  as the initial set of adopters remains constant.

\textit{Transition rule:} The transition rule consists of the following components: nodes (household ids), the from and to state of transition (0 to 1), the rule name, the node threshold, the node probability, the personal benefit, and the weights associated with personal, community and neighbor components.  The node threshold specifies the barriers for a household to adopt solar as outlined in the literature~\cite{reames2020distributional}. It includes factors such as internet access, language proficiency, racial and socioeconomic status, rental housing, education level, income, house age, and the age of household members. The node threshold value is calculated by dividing the sum of barriers using a step function that ranges from 0.1 to 0.95. We exclude the boundary values of 0 and 1 to avoid extreme cases where a household would be either forced to adopt or prevented from transitioning to an adopter. Node probability captures the likelihood of adopting solar, influenced by policies and incentives. It reflects the impact of external interventions, allowing for a constant node probability to signify no interventions or a variable node probability to indicate changes influencing over time. This flexibility enables the model to simulate the dynamic nature of solar adoption under various policy and incentive conditions. 

The different cases, along with the node probability values, are described below:

Case 1a and Case 1b: Maintain a consistent node probability across all groups, set at 0.1 and 0.2, respectively. 

Case 2a and Case 2b: We assign a higher node probability to LMI households. This approach is designed to introduce specialized incentives for this demographic, resulting in a higher node probability within the LMI community. Specifically, we assign a node probability of 0.1 to all non-LMI households and 0.2 and 0.5 to LMI households for Case 2a and Case 2b, respectively.

Case 3: We derive node probabilities by emulating the adoption pattern in the LMI community as described in the paper by O’Shaughnessy et al.~\cite{o2021impact}. The authors note an increase in adoption rates in the LMI community following the introduction of a policy, followed by a decrease in subsequent time steps and a steady rise after that. We model this phenomenon by initially introducing a higher node probability for the LMI community in the first time step. This probability is then adjusted to be on par with the non-LMI community, and a gradual increase is introduced later in the following steps.

Case 4: This case is inspired by the 30\% solar tax credit offered by the Federal government~\cite{solar_rebate}. In this approach, we utilize data on the potential solar energy production for all households, applying the method outlined in Subsection~\ref{subsec:generatin}. We then calculate the mean installation cost based on Virginia's current average installation cost per watt (\$3.04)~\cite{solar_cost}. The households are subsequently sorted according to the rebate they receive. They are categorized into bins with values ranging from 0 to 1 for the node probability.

Case 5: This case closely resembles Case 4, with the key distinction being that the Low-Moderate Income (LMI) community receives an additional 20\% solar tax credit~\cite{lmi_solar_cost}. The methodology for generating node probability remains the same as in the previous case. 

Personal benefit is derived from the average solar energy generation. Our proposed model, described for generating PV profiles in Subsection~\ref{subsec:generatin}, is used for this purpose. The values are then normalized to [0,1]. The community adoption rate is the ratio of solar-adopting households to total households in a county, and the neighborhood adoption rate counts immediate solar-adopting neighbors. The values for these parameters are obtained during the simulation.
We assign weights of 0.4, 0.3, and 0.3 to personal benefit value, community influence, and neighbor influence, respectively, giving slightly more emphasis to the personal component than the other two factors. The sum of weights is equal to 1.

The list of modeling parameters along with a short description, range, and values chosen is shown in Table~\ref{tab:parameters}. Parameter values are either obtained from the literature or calibrated during the simulation setup. 

%% file: appendix.tex
%%%%
\renewcommand{\figurename}{Supplementary Figure}
\renewcommand{\tablename}{Supplementary Table}

\newcommand{\algofontsize}{\footnotesize}

\renewcommand{\thesubsection}{Supplementary Note \arabic{subsection}}
%\renewcommand{\thesection}{S\arabic{section}}

% \linespread{1.05} % Adjust line spacing to achieve 1.05 spacing.
\raggedbottom
%%\unnumbered% uncomment this for unnumbered level heads

\fontsize{11}{12}\selectfont %{a}{b}: a: font size, b: line spacing

\section*{Supplementary information}

\subsection{: Datasets, variables and explanations}
\label{sec:apendix_dataset}

The different datasets used in this research, along with the sections where they are used, is mentioned in Table~\ref{tab:datasets}.

\begin{table}[ht]
\centering
\caption{Datasets used for this research}
\label{tab:datasets}
\begin{tabular}{|p{8cm}|p{5cm}|}
\hline
\textbf{Dataset} & \textbf{Usage section} \\ \hline
US Census PUMS~\cite{pums} & Methodology \\ 
Residential Energy Consumption Survey (RECS 2020) v2~\cite{RECS} & Methodology \\
Synthetic population~\cite{adiga15US} & Methodology \\
National Solar Radiation Database (NSRDB)~\cite{nsrdb} & Methodology \\
Climate Regions by County~\cite{antonopoulos2022guide} & Methodology \\
LBNL~\cite{barbose2021residential} & Methodology, Validation \\
Pecan Street~\cite{pecan_street} & Validation \\
DeepSolar~\cite{wussow2024exploring} & Validation \\
\hline
\end{tabular}
\end{table}

\begin{table}[ht]
\centering
\caption{ Variables used and their explanations}
\begin{tabular}
{|p{4cm}|p{8cm}|}
\hline
\textbf{Variables} & \textbf{Explanations} \\
\hline
NHSLDMEM & Number of household members \\
BEDROOMS & Number of bedrooms \\
TYPEHUQ & Type of housing unit \\
FUELHEAT & Main space heating fuel \\
KOWNRENT & Ownership status of the unit \\
YEARMADERANGE & Year range in which the housing unit was built \\
MONEYPY & Annual household income range \\
BA\_climate & Building's climate zone \\
SQFTRANGE & Square footage range of the household \\
SOLAR & Rooftop solar adoption\\
PV & Photovoltaic\\
ML & Machine learning \\
XAI & Explainable artificial intelligence \\
LBNL & Lawrence Berkeley Lab\\
LMI & Low-to-Moderate-Income\\
RECS & Residential Energy Consumption Survey\\
NREL & National Renewable Energy Laboratory\\
JSD & Jensen Shannon Divergence\\
KDE & Kernel Density Estimation \\
SMOTEN & Synthetic Minority Over-sampling Technique for Nominal \\
SVM & Support Vector Machine\\
GPR & Gaussian Process Regressor\\
EI & Expected Improvement \\
CDF & Cumulative Distribution Function \\ 
PDF & Probability Density Function \\
\hline
\end{tabular}
\label{tab:variables_abbreviations}
\end{table}

\subsection{: Solar adoption pre-processing step: SQFTRANGE classification}
\label{sec:apendixA}

\begin{figure}[!h]
    \centering
    \includegraphics[width=0.99\textwidth]{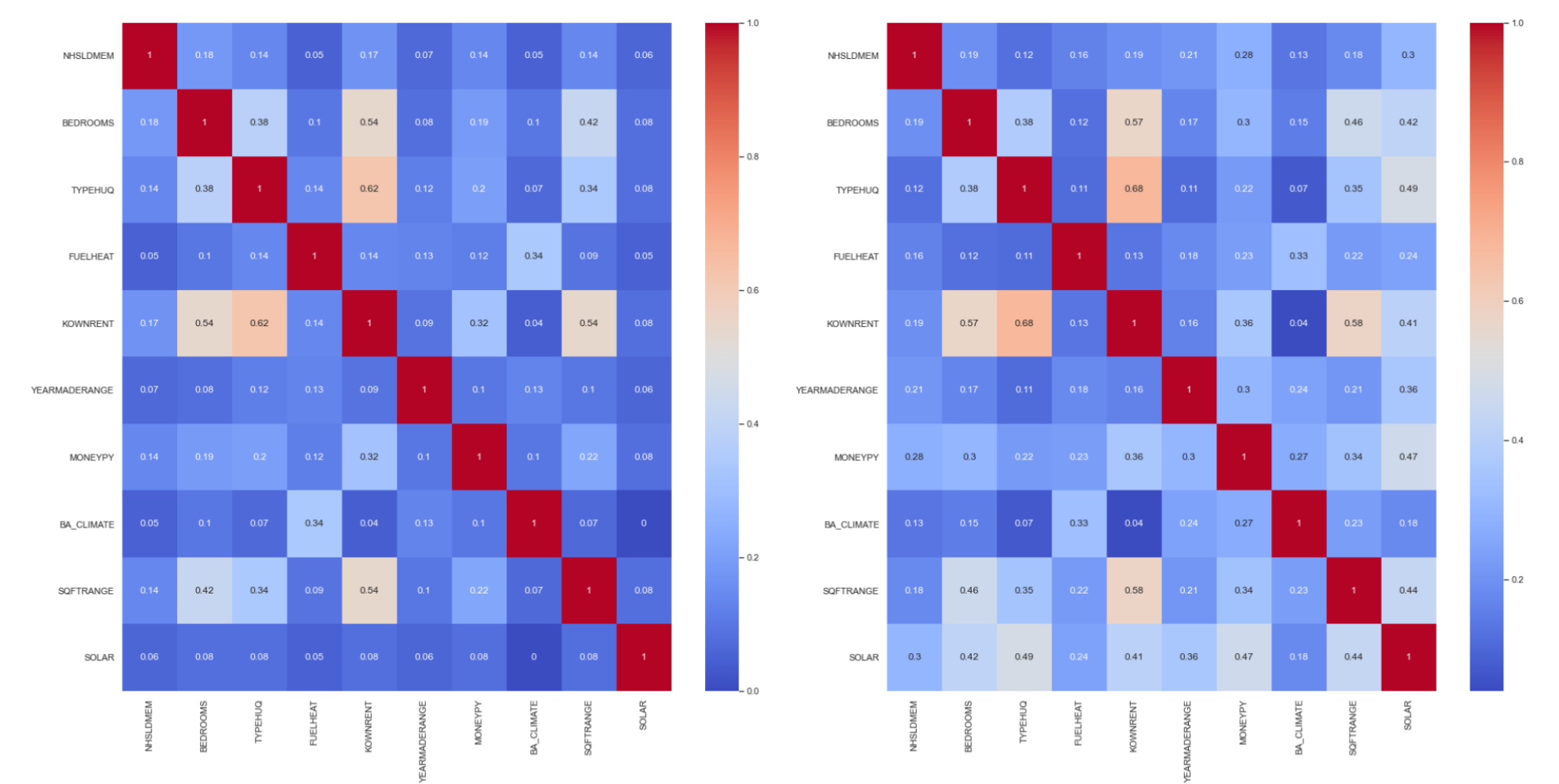}
    \caption{Cramer's V correlation matrix for the South Atlantic division is depicted in this figure, providing insights into the correlation between all input features and the output feature used for classifying square footage range. The left panel of the figure displays the correlation matrix of the original dataset from the South Atlantic division of the RECS survey, whereas the right panel presents the correlation matrix after applying the oversampling technique, SMOTE. A high correlation is indicated by a value of 1 and a hue color of red, while a low correlation between features is shown in blue, indicated by a value of 0. The correlation between features remains consistent after the application of SMOTE.}
    \label{fig:sa_correlation_matrix}
\end{figure}

SQFTRANGE classification is a pre-processing step before proceeding with solar adoption, as this is one of the features used to predict adoption. The correlation matrix before and after SMOTE to address the class imbalance issue in the South Atlantic division is presented in Figure~\ref{fig:sa_correlation_matrix}. The values in the matrix range from 0, indicating low correlation, to 1, denoting high correlation. The left panel displays the matrix for the original dataset from the South Atlantic division of the RECS survey, while the right panel shows the matrix post-application of SMOTE. The correlation between features remains consistent after applying SMOTE. This consistency is observed in other divisions as well. Besides input features, the matrix also includes the output label SQFTRANGE. Furthermore, the variable `SOLAR' is included to examine its correlation with other variables. 

Next, the square footage range classification performance metrics are presented in Table~\ref{tab:performance_metrics}. Although the accuracy and F1 scores for all divisions fall below 80\%, the confusion matrices in Table~\ref{tab:confusion_matrix_1} for the validation set and Table~\ref{tab:confusion_matrix_2} for the test set indicate that the square footage range is predominantly classified into neighboring ranges, which is acceptable for our analysis. The other divisions exhibit patterns similar to those of the South Atlantic division.

\begin{table}[h]
\caption{ Performance Metrics for each of the five chosen divisions: The table consists of Test Accuracy percentages from the RF classifier, GB classifier, and SVM classifier presented in the second column, and F1 score metrics in the percentage provided in the third column.}\label{tab:performance_metrics}
\begin{tabular*}{\textwidth}{@{\extracolsep\fill}lcccccc}
\toprule
& \multicolumn{3}{c}{Accuracy} & \multicolumn{3}{c}{F1 Score} \\
\cmidrule{2-4}\cmidrule{5-7}
Division & RF & GB & SVM & RF & GB & SVM \\
\midrule
South Atlantic Division & 69& 69& 64& 67& 67& 65\\
Pacific & 78& 74& 61& 78& 74& 64\\
West South Central & 58& 58& 58& 57& 57& 59\\
Mountain North & 55& 55& 48& 54& 53& 50\\
New England & 68& 62& 58& 70& 62& 69\\
\bottomrule
\end{tabular*}
\end{table}
% First Confusion Matrix
\begin{table}[h]
\caption{ Confusion matrix for validation set for South Atlantic division}\label{tab:confusion_matrix_1}
\centering
\begin{tabular}{cccccccc}
\toprule
141 & 0 & 5 & 8 & 0 & 0 & 0 & 0 \\
0 & 136 & 3 & 20 & 2 & 1 & 0 & 0 \\
1 & 13 & 104 & 47 & 2 & 0 & 0 & 0 \\
1 & 4 & 6 & 88 & 20 & 2 & 1 & 5 \\
0 & 0 & 3 & 61 & 55 & 30 & 9 & 5 \\
0 & 0 & 1 & 32 & 18 & 72 & 16 & 16 \\
0 & 0 & 0 & 21 & 7 & 19 & 110 & 23 \\
0 & 0 & 0 & 12 & 6 & 14 & 15 & 105 \\
\bottomrule
\end{tabular}
\end{table}

% Second Confusion Matrix
\begin{table}[h]
\caption{ Confusion matrix for test set for South Atlantic division}\label{tab:confusion_matrix_2}
\centering
\begin{tabular}{cccccccc}
\toprule
14 & 0 & 0 & 0 & 0 & 0 & 0 & 0 \\
0 & 14 & 0 & 2 & 0 & 0 & 0 & 0 \\
0 & 1 & 18 & 7 & 1 & 0 & 0 & 0 \\
0 & 0 & 2 & 11 & 3 & 1 & 0 & 0 \\
0 & 0 & 0 & 6 & 5 & 2 & 2 & 0 \\
0 & 0 & 1 & 5 & 2 & 4 & 3 & 1 \\
0 & 0 & 0 & 1 & 0 & 3 & 11 & 1 \\
0 & 0 & 0 & 3 & 0 & 4 & 0 & 12 \\
\bottomrule
\end{tabular}
\end{table}

\begin{figure}[!h]
    \centering
    \includegraphics[width=0.95\textwidth]{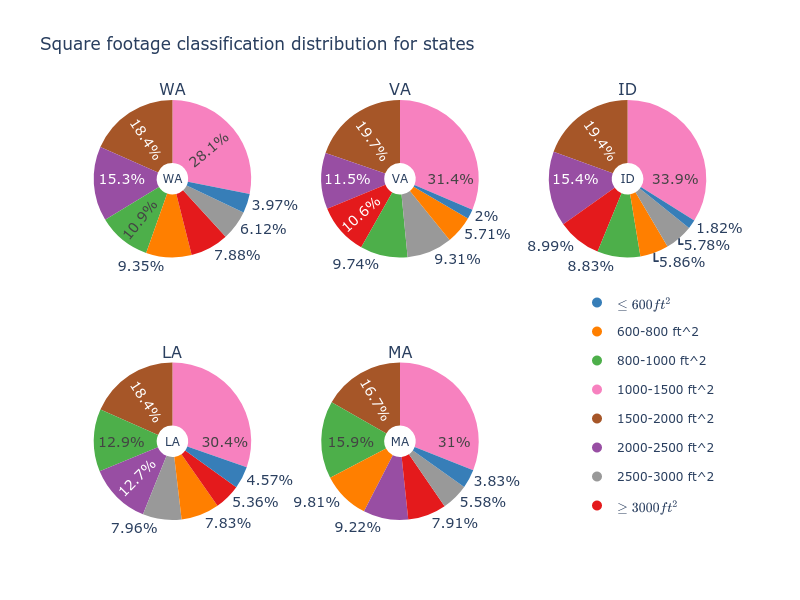}
    \caption{\textbf{ SQFTRANGE classification of the representative states WA, VA, ID, LA, and MA.} This is predicted by the trained ensemble model and the synthetic population's voting mechanisms. SQFTRANGE from 1000 $ft^2$ to 2000 $ft^2$ constitute nearly half of the distribution in these states (51.1\%, 48.8\%, 53.3\%, 46.5\%, and 47.7\% respectively). 
    }
    \label{fig:sqft_distribution_pie}
\end{figure}

The SQFTRANGE classification predicted by the trained ensemble model, along with the voting mechanisms in the synthetic population of the representative states in Table~\ref{tab:census}, is depicted in Figure~\ref{fig:sqft_distribution_pie}. 

SQFTRANGE follows the RECS class range. SQFTRANGE from 1000 $ft^2$ to 2000 $ft^2$ constitutes nearly half of the distribution in all the five chosen states. SQFTRANGE, less than 600 $ft^2$, appears as the least frequent in the distribution. 

\subsection{: Solar adoption performance matrix}
\label{sec:apendixA1}

$\beta$ value to reduce wrong predictions is set between 0 and 2. We also adjust the decision threshold, $\tau$, from 0.05 to 0.95 to further address the class imbalance issue. The values of $\beta$ and $\tau$, along with accuracy and F1-score, are shown in Table~\ref{tab:performance_metrics_solar}. Most states showed high accuracy and F1-score, but MA had lower scores, around 70\%. The difference between the actual number of solar adopters and our predicted number is small, at 1.95\%, according to the Berkeley Lab dataset~\cite{barbose2021residential}.

\begin{table}[h]
\caption{ Performance metrics for solar adoption classification in each of the five chosen states.} 
%\swapna{Fix punctuation. Cite Table VXYZ.} - Done}
\label{tab:performance_metrics_solar}
\begin{tabular*}{\textwidth}{@{\extracolsep\fill}lcccc}
\toprule
{States } & {$\beta$} & {$\tau$} & {Accuracy} & {F1-score}\\
\midrule
VA&1.47 &0.53&97\%&97\%\\
WA&0.56&0.67&99\%&99\%\\
LA&0.62&0.32&96\%&96\%\\
ID&1.29&0.37&99\%&99\%\\
MA&0.9&0.36&71\%&68\%\\
\bottomrule
\end{tabular*}
\end{table}

\subsection{: More insights on Spatial and temporal solar production results}
\label{sec:apendixB}

\begin{figure}[!h]
    \centering
    \includegraphics[width=0.95\textwidth]{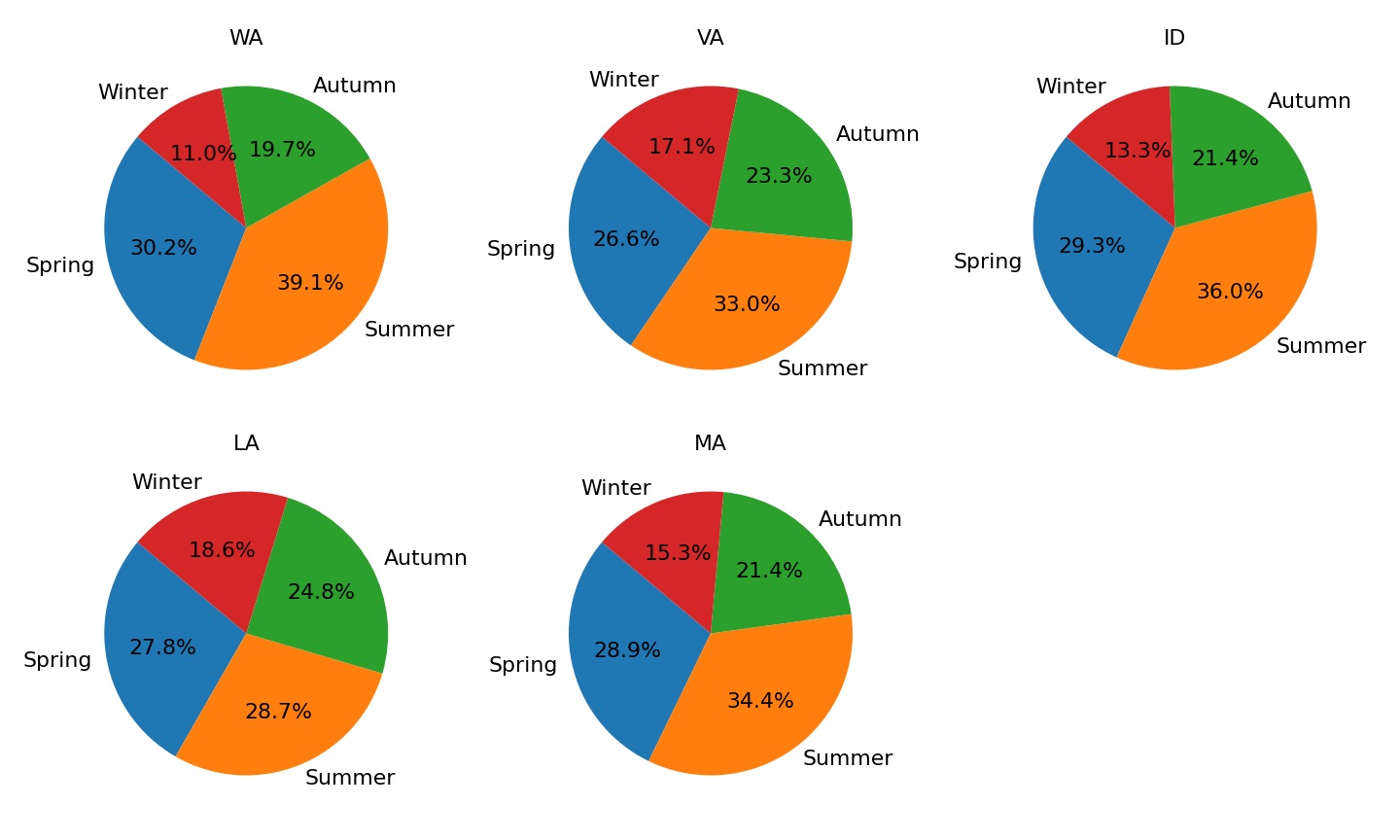}
    \caption{\textbf{ Seasonal solar energy production for WA, VA, ID, LA, and MA.}~Notably, during winter, WA and ID contribute 11\% and 13.3\%, respectively, to solar energy generation. MA follows with a 15.3\% contribution in winter. All three states — WA, ID, and MA —share a common cold climate zone, which influences their solar energy production. Additionally, WA's marine climate zone results in increased cloud cover, further impacting the amount of sunlight received and, thus, solar energy generation.} 
    \label{fig:seasonal_variability}
\end{figure}

Pie charts in Figure~\ref{fig:seasonal_variability} reveal seasonal solar production variations, highlighting that summer has the highest output across all states, followed by spring, autumn, and winter. WA and ID contribute 11\% and 13.3\% in winter, while MA contributes 15.3\%, indicating that cold climates impact solar output. LA shows consistent summer, spring, and fall contributions around 25\% to 29\%, with a notable 19\% in winter.

\begin{figure}[!h]
    \centering
    \includegraphics[width=0.95\textwidth]{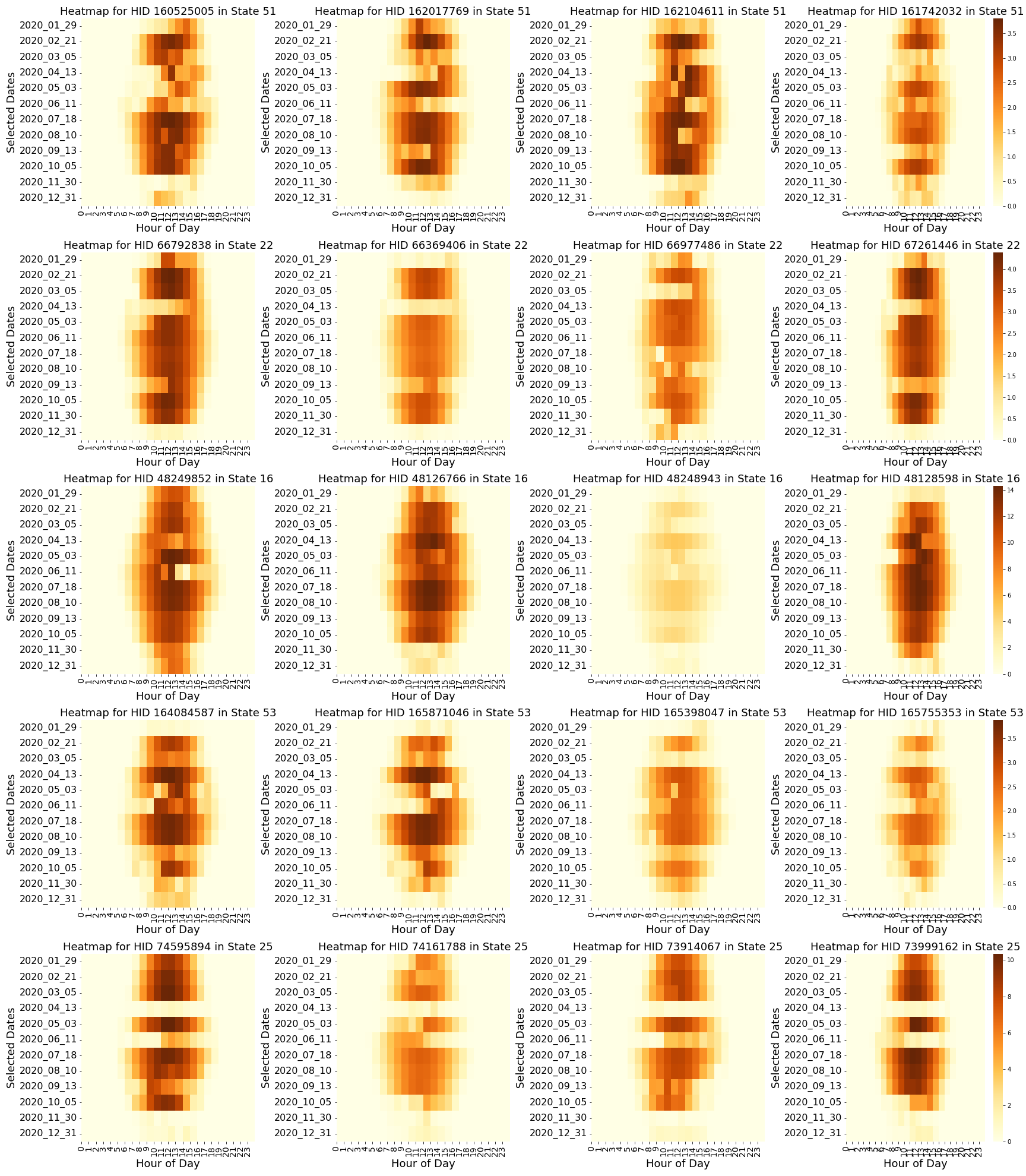}
    \caption{\textbf{ Comparative heatmap analysis of solar energy production across five states.} This multi-row heatmap presents a state-by-state comparison of solar energy generation in randomly selected households. Each row corresponds to a different state, with the first row representing VA, LA, ID, WA, and MA. In each state's row, four representative households have been selected to demonstrate spatial similarities and differences in solar energy production. The x-axis of each heatmap row represents the hour of the day, ranging from 0 to 23, while the y-axis comprises randomly selected dates from each month, totaling 12 dates to illustrate each month's solar energy patterns. The heatmap provides a visual representation of temporal and spatial variability in solar energy generation across diverse geographic regions. Each state's row reveals unique trends and anomalies in solar production, reflective of local climatic conditions and solar potential.} 
    \label{fig:heatmap}
\end{figure}

The heatmap in Figure~\ref{fig:heatmap} shows daily solar energy patterns for four homes in each of the five states, chosen randomly from each month. In VA, homes are from Fairfax, Hampton, Lynchburg, and Wythe counties, covering a range of locations and sizes (1100-1700 ft²). All homes see lower solar production in November and December. An interesting April pattern shows the first two eastern homes with nearly no production at times, unlike the central and western homes. This possibly denotes cloud cover effects. Similar patterns are observed in LA and MA, suggesting regional cloud impacts. In December, an eastern home in VA shows minimal solar production, emphasizing climate's role in solar generation. For LA, the study includes homes from East Carol, Beenville, Jefferson, and Morehouse, varying in size (1600-2700 ft²). Homes in neighboring eastern counties and a distinct southern county's home show different solar production patterns, with December seeing minimal production in most homes, highlighting seasonal effects. In ID, homes from Butte, Adams, Boundary, and Bannock reflect diverse solar production patterns, with square footage ranging from 1200-2800 ft². Notably, a northern home (Boundary) shows unique patterns, and a localized cloud event in June significantly affected solar production. WA data for homes in Clark, Snohomish, Pacific, and Skagit counties (900-2300 ft²) reveal varied production trends, even among neighboring counties. Coastal homes exhibit unique patterns, with minimal production in January, consistent with seasonal variability observed in Figure~\ref{fig:seasonal_variability}. MA covers homes in Hampden, Dukes, Berkshire, and Bristol, varying widely in size (800-3000 ft²). January shows unusually high production compared to other states, while November and December are low. A specific June date shows no morning production in the first and third homes, with Dukes County's home maintaining normal levels but averaging lower production overall, indicating micro-geographical impacts on solar output.

\begin{figure}[!h]
    \centering
    \includegraphics[width=0.95\textwidth]{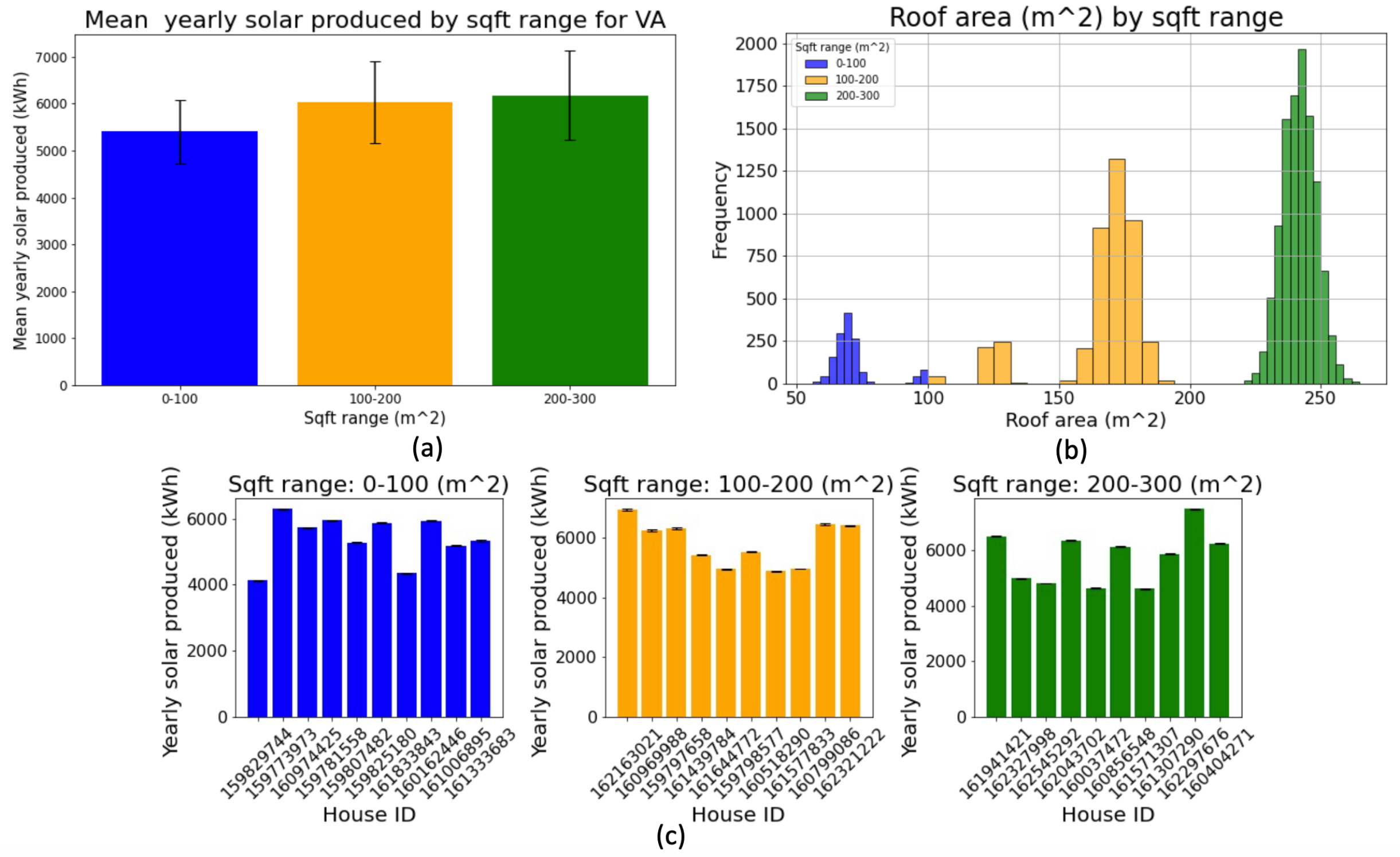}
    \caption{\textbf{ Analysis of yearly aggregate residential PV solar production in VA by roof area in square meters ($m^2$)}. (a) A bar chart shows average yearly solar production and standard deviations across different roof sizes, illustrating the efficiency relation between property size and solar output. (b) A histogram indicates the distribution of households by roof size in VA, providing context for the production data. (c) A detailed view of yearly solar production for ten randomly selected households from each size category shows individual variability and the diverse potential for solar generation among properties.} %\swapna{Too much detail. Shorten the explanation to highlight key points only.}}-Done
    \label{fig:va_sqft_solar}
\end{figure}

Finally, we analyze the yearly aggregate residential solar production with respect to roof area in the state of VA, as depicted in Figure~\ref{fig:va_sqft_solar}. This figure comprises three plots, each offering a different perspective on solar production relative to property size. The first plot provides a comprehensive overview of average yearly solar production, including its uncertainties, across various roof areas. The second plot delves into detail, presenting the distribution of households. Here, the frequency distribution is notably higher for roof areas ranging between 200-300 $m^2$ compared to other ranges. The plots in the second row further enhance this analysis by providing insights into the average solar production and its uncertainties across ten randomly selected households in each roof area range. Across all three roof area categories, the annual average solar production shows slight variations, mostly ranging between 4000 kWh and 7000 kWh. While total roof area is a contributing factor to this variability, other elements such as the number of planes, tilt, and orientation of the panels, as well as the location of the households also play a significant role in solar energy production.

\begin{figure}[!h]
    \centering
    \includegraphics[width=0.95\textwidth]{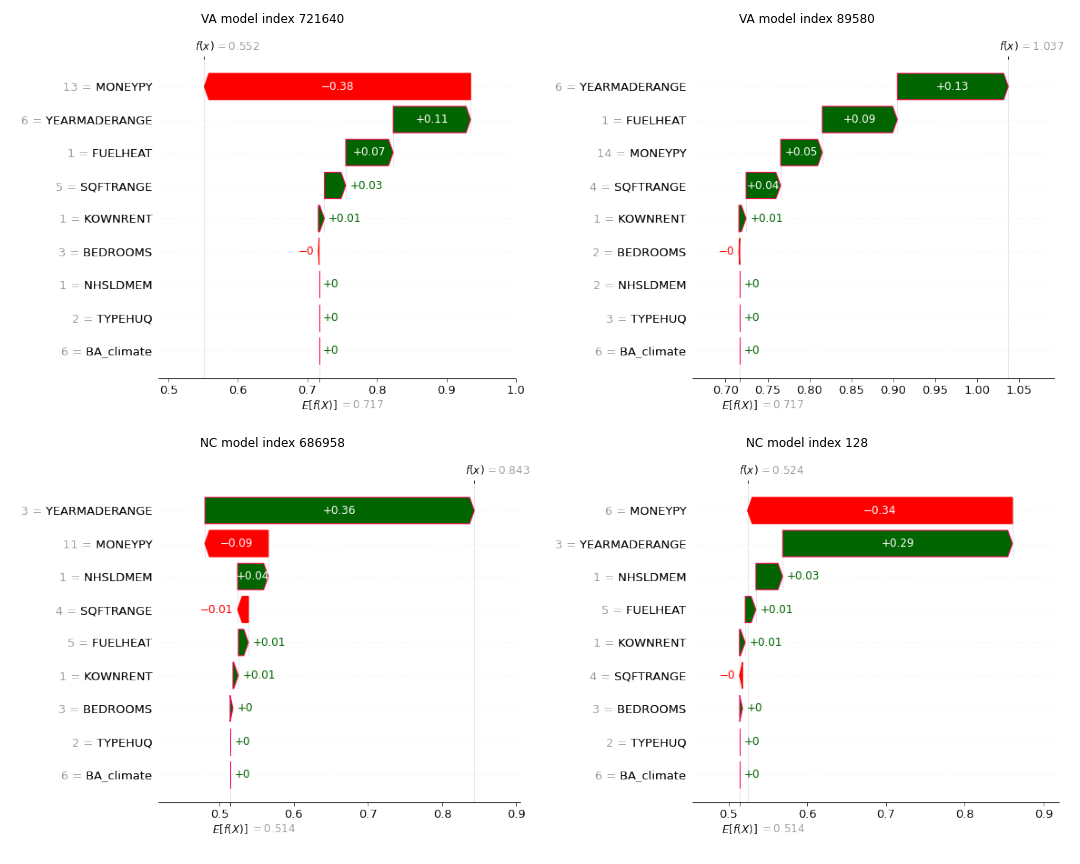}
    \caption{\textbf{ Local SHAP importance plot for solar Adopters in VA and NC}: This figure shows SHAP value contributions for four households in VA and NC. In VA, one household's features all positively impact the SHAP value, while another's MONEYPY feature significantly lowers its prediction probability from over 0.9 to 0.552. The mean value ($E(f(x))$) benchmarks these contributions, with green bars indicating positive effects and red bars showing negative ones. In NC, one household sees a major drop in prediction probability to 0.524 due to MONEYPY, whereas another's YEARMADERANGE greatly boosts its probability to 0.843.}
    \label{fig:xai_va_nc_2}
\end{figure}

\subsection{: More insights on XAI of state-level models}
\label{sec:apendixC}

\begin{figure}[!h]
    \centering
    \includegraphics[width=0.99\textwidth]{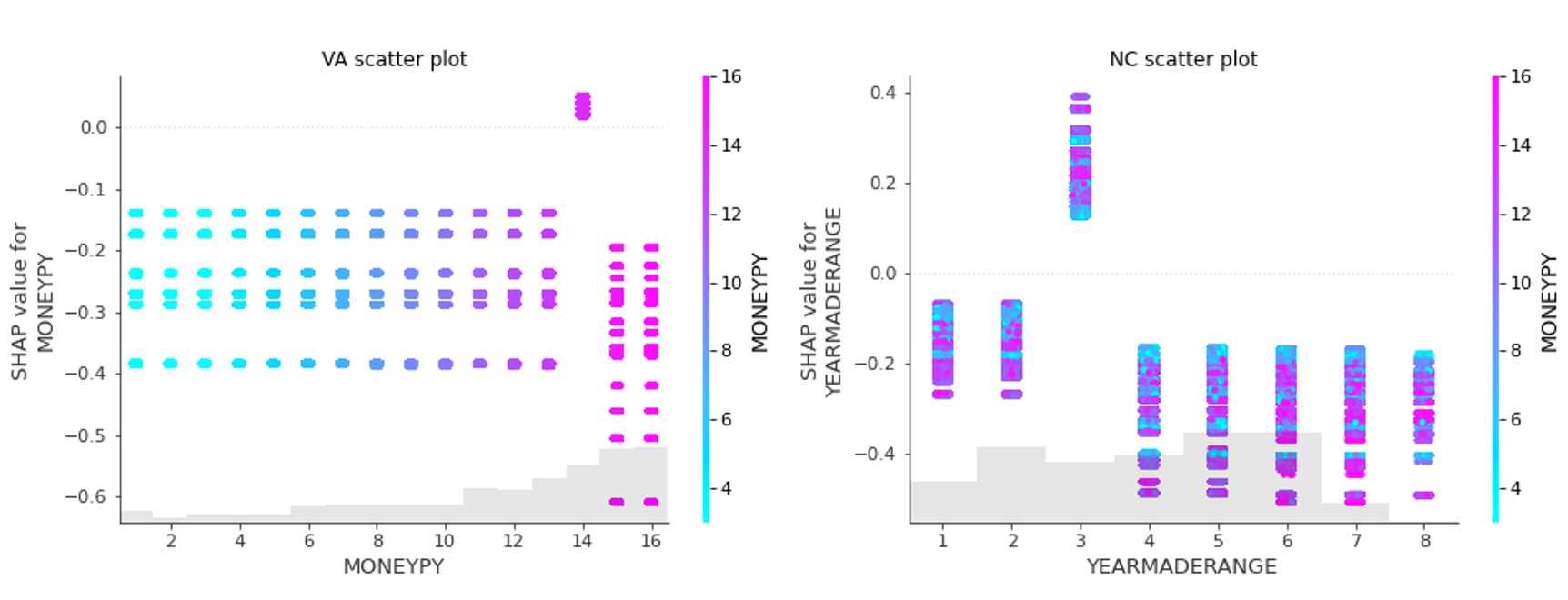}
    \caption{ VA model scatter plot demonstrates the MONEYPY distribution along with its SHAP contribution in VA. The grey histogram represents the data distribution for MONEYPY. A MONEYPY value of 14 contributes positively to the SHAP value, whereas all other values contribute negatively. Notably, MONEYPY values of 15 and 16 have a stronger negative impact on SHAP values. The variance in SHAP values for each MONEYPY value is attributed to the presence of feature interactions. NC model scatter plot illustrates the feature interaction between YEARMADERANGE and MONEYPY in North Carolina's SHAP contributions. Although both households in NC have a YEARMADERANGE value of 3, household 1 shows a +0.26 positive contribution, and household 2 shows a +0.36 positive contribution. The scatter plot reveals that YEARMADERANGE generally contributes positively when compared to other values within the same range. However, a higher MONEYPY value (represented in pink) further increases the SHAP value compared to lower values (shown in blue), demonstrating the significant impact of feature interaction on SHAP values.}
    \label{fig:xai_va_nc_3}
\end{figure}
Figure~\ref{fig:xai_va_nc_2} delves into individual SHAP waterfall plots for households in VA and NC, showcasing how different features affect solar adoption predictions. These plots highlight the importance of diversity in features across households. 

\begin{figure}[!h]
    \centering
    \includegraphics[width=0.95\textwidth]{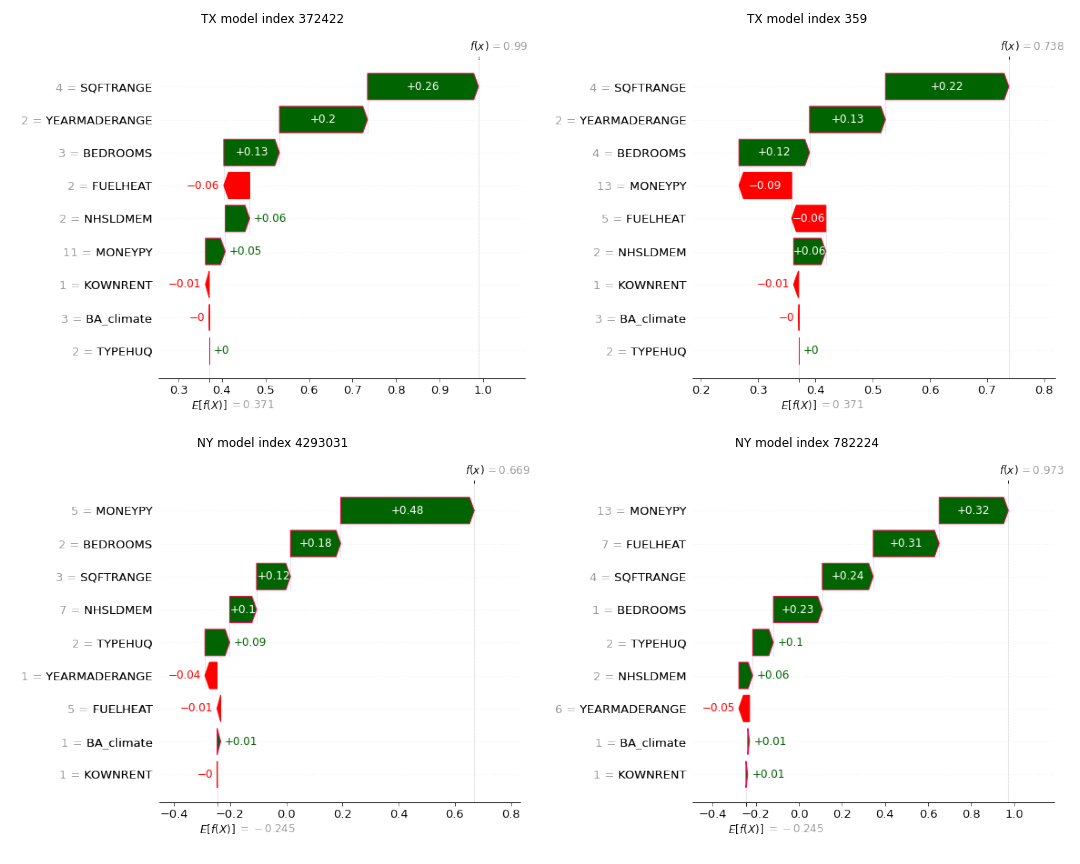}
    \caption{\textbf{ Household-level (Local) SHAP importance plot}: The figure displays SHAP values for households in TX and NY, showing how various features impact the model's solar adoption predictions. Green bars in the plots indicate positive contributions to the prediction, while red bars show negative influences. The mean model output is marked by $E(f(x))$, and the final output for each household is denoted by $f(x)$. In TX, property size significantly affects predictions for the chosen households, suggesting a strong link between square footage and solar adoption. In NY, household income is a key factor, indicating its critical role in predicting solar adoption. All plots reveal that the final model outputs for these households exceed their state's average, highlighting the importance of these features in solar adoption.}
    \label{fig:xai_tx_ny_2}
\end{figure}
Figure~\ref{fig:xai_va_nc_3} showcases two scatter plots, with the first pertaining to VA and the second to NC. In the VA plot, the focus is on the variability of a single feature, while the NC plot delves into the interplay between two key features: household income and year of construction. In the NC plot, YEARMADERANGE typically yields a positive SHAP value contribution in comparison to other values in its range. This effect is further amplified when paired with higher MONEYPY values (indicated by the pink coloration), in contrast to lower values (represented in blue), highlighting the significant influence of feature interaction on SHAP values.

Similar to the VA-NC model, Figure~\ref{fig:xai_tx_ny_2} presents household-level plots to provide local insights into feature contributions. Unlike the household-level plots in VA-NC, where the same feature shows high importance across households within the state, the importance varies across both TX and NY. All four plots demonstrate that the final model output $f(x)$ for these households is higher than the state's mean output $E(f(x))$. Additionally, most feature values exhibit either a positive or negligibly negative impact on the model. 

\begin{figure}[!h]
    \centering
    \includegraphics[width=0.95\textwidth]{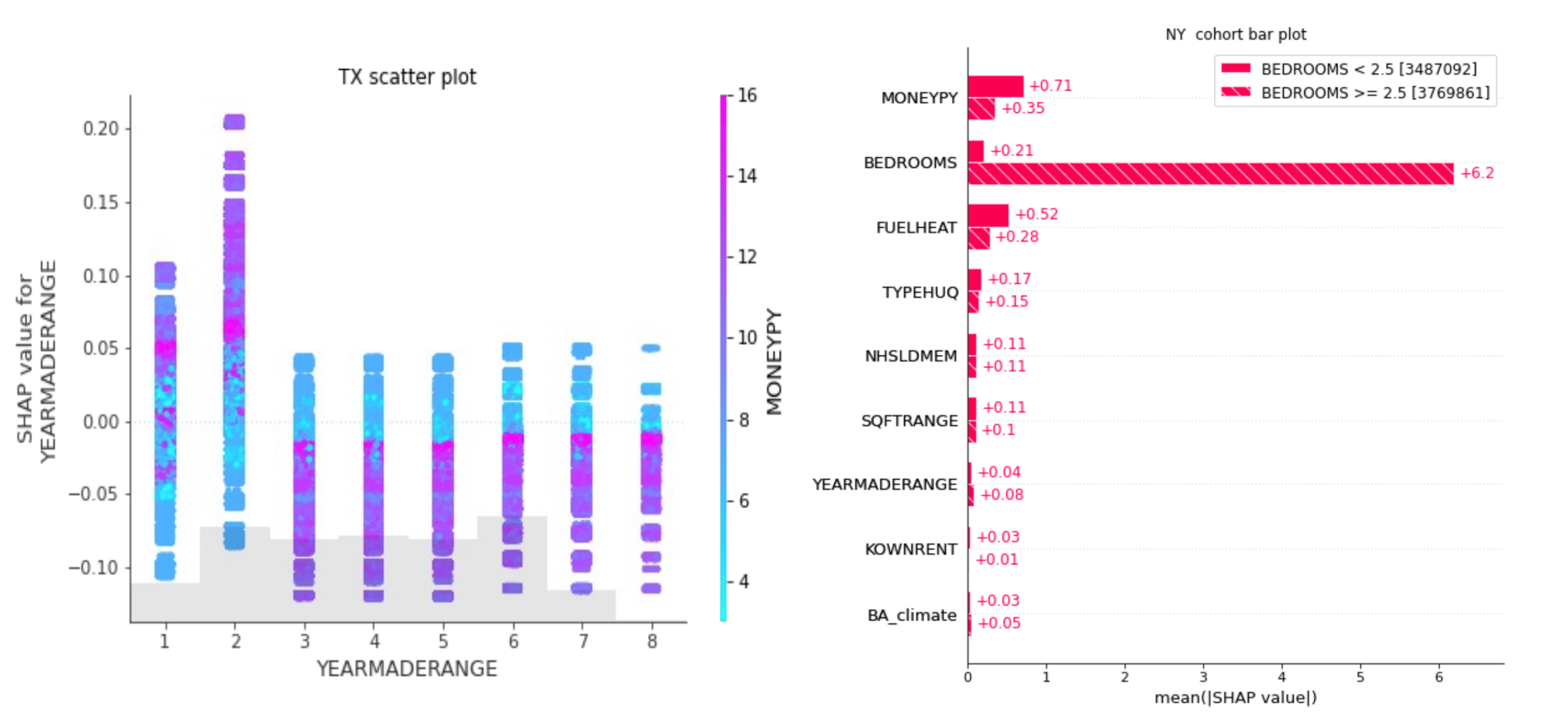}
    \caption{ The scatter plot for the TX model illustrates the feature interaction between YEARMADERANGE and MONEYPY in relation to Texas's SHAP contributions. Lower-income values are represented in light blue, while higher-income values are depicted in pink. The SHAP values are positioned along the y-axis, starting with the lowest values at the bottom and increasing towards the top. It is evident that lower values, particularly a value of 2 in YEARMADERANGE, yield higher SHAP values. In this context, lower-income values exhibit lower SHAP values, whereas middle and high-income values correlate with higher SHAP values. However, distinguishing between the SHAP contributions of high and medium income is challenging, as the pair-wise feature interaction does not provide sufficient differentiation. The cohort multi-bar plot for the NY model displays cohorts created based on the feature BEDROOMS, with one cohort comprising households with BEDROOMS $<$ 2.5 and the other with BEDROOMS $\ge$ 2.5. Each bar in the plot represents a separate cohort within the multi-bar layout. This arrangement is utilized to provide a global summary of feature importance, allowing for optimal separation of the SHAP values of the instances. Additionally, it is worth noting that the mean SHAP values are exceptionally high, around 6.2, for the cohort with BEDROOMS $\ge$ 2.5. }
    \label{fig:xai_tx_ny_3}
\end{figure}

Figure~\ref{fig:xai_tx_ny_3} is a scatter plot for Texas, which facilitates the understanding of the significance of interactions between features. It clearly shows that lower values, particularly a value of 2 in YEARMADERANGE, are associated with higher SHAP values. In this scenario, lower income values result in lower SHAP values, while middle and high-income values are correlated with higher SHAP values. However, it remains challenging to differentiate between the SHAP contributions of high and medium income levels, as the pairwise feature interaction does not offer adequate distinction. Furthermore, the SHAP value is significantly lower for YEARMADERANGE values greater than 2. The bars (composed of points) appear flipped in these ranges, indicating that the SHAP value increases as the MONEYPY decreases in these ranges of YEARMADERANGE.

The last plot in the series for the TX-NY model comparison is a cohort multi-bar plot for the NY model, as shown in Figure~\ref{fig:xai_tx_ny_3}. This plot categorizes the data into different cohorts based on a specific feature and then compares the mean SHAP values of these cohorts using a multi-bar plot. This form of visualization aids in understanding the impact of features on the model's predictions with respect to a particular segment. It is important to note that actual SHAP values can be either positive or negative, while the cohort bar plot primarily indicates the significance of a particular feature in influencing the SHAP value. In this figure, the cohorts are formed based on the key feature `BEDROOMS'. While BEDROOMS $\ge$ 2.5 shows a higher feature contribution with a SHAP value of 6.2, it is crucial to recognize that the direction of this contribution is negative. This insight is obtained by analyzing this plot in conjunction with the bee swarm plot for the NY model presented in Figure~\ref{fig:explainability}(b).

\subsection{: Comparison of solar adoption between datasets}
\label{sec:apendixV}

%(a) Stacked bar comparison of solar adopter counts, where the lower and upper bounds are established by either LBNL or DeepSolar counts. The bars are pink when the synthetic adopter count is below both datasets. The bars are brown if the synthetic adopter count is above both datasets. 

\begin{figure}[!h]
    \centering
    \includegraphics[width=1\textwidth]{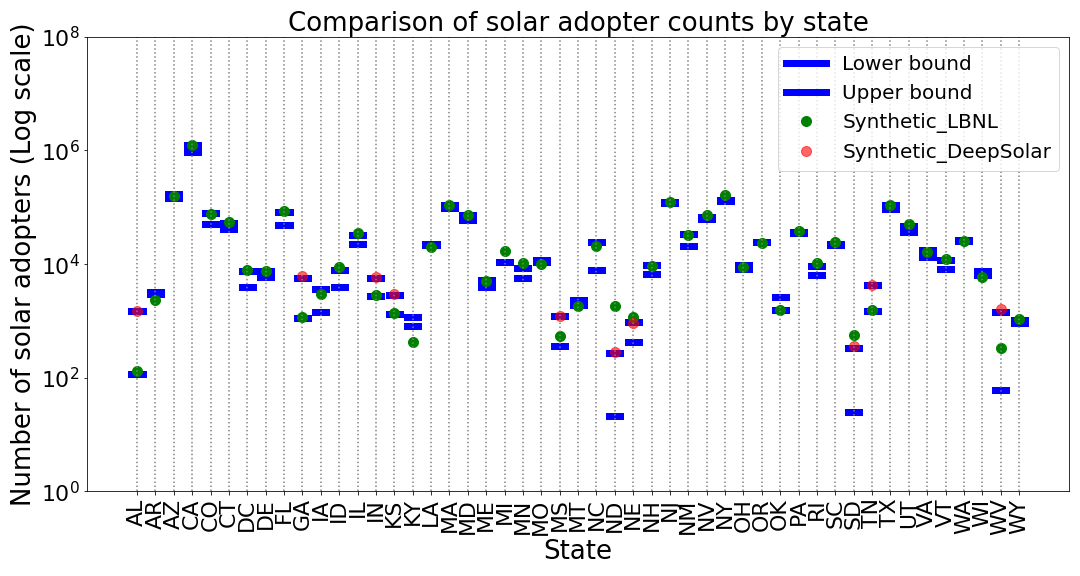}
    \caption{\textbf{ Comparison of different solar adopter datasets across U.S. states.} The x-axis represents the contiguous states in the U.S., and the y-axis represents the number of solar adopters on a log scale. The lower and upper bounds are specified by either LBNL or DeepSolar counts. The green and red points represent the synthetic solar adopter counts calibrated using LBNL and DeepSolar solar adopter counts as the ground truth, respectively. Note that the calibration with DeepSolar is performed in ten states.} 
    \label{fig:comparison_adopters}
\end{figure}

\begin{figure}[!h]
    \centering
    \includegraphics[width=1\textwidth]{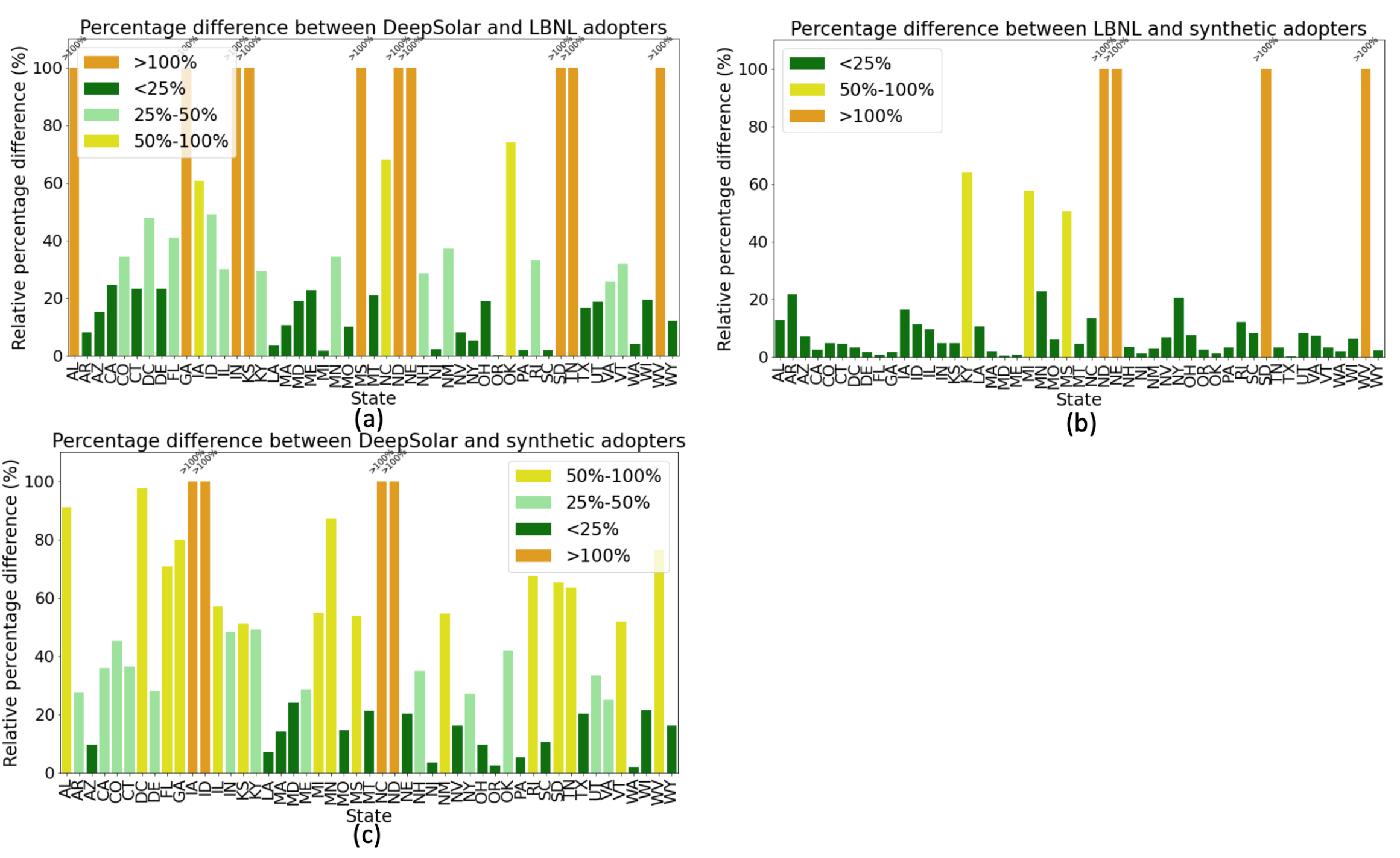}
    \caption{\textbf{ Relative percentage comparison of solar adopter datasets} (a) Relative percentage comparison of real-world solar adopter data sets, LBNL, and DeepSolar across U.S. states. (b) Relative percentage comparison of LBNL and synthetic solar adopter datasets. (c) Relative percentage comparison of DeepSolar and synthetic solar adopter datasets. In a,b, and c, the x-axis represents the contiguous states in the U.S., while the y-axis denotes the relative percentage difference between real-world data and synthetic data for solar adopters. In cases where the percentage difference exceeds 100\% (orange bars), an annotation is added at the cap level. It is important to note that there were no states with a percentage difference between 25\% and 50\% between the LBNL and synthetic adopter datasets.} 
    \label{fig:validation_adopters}
\end{figure}

We compared different real-world datasets (LBNL and DeepSolar) with our synthetic solar adopter dataset in Figure~\ref{fig:comparison_adopters}. LBNL and DeepSolar define the lower and the upper bound in each state. They are calculated as:
\begin{align}
  \text{lower\_bound} &= \min(\text{lbnl\_adopter}, \text{deepsolar\_adopter}) \\
  \text{upper\_bound} &= \max(\text{lbnl\_adopter}, \text{deepsolar\_adopter})
\end{align}
 
Synthetic solar adopters calibrated with LBNL dataset and DeepSolar dataset as the ground truths are depicted as green and red points respectively. The synthetic solar adopter counts are comparable to either of the datasets. Next, we compared the two available datasets in Figure~\ref{fig:validation_adopters}a. In 10 states, the relative percentage exceeds 100\%, revealing significant differences between the two datasets. It is important to note that the LBNL solar adopter count is based on 2020 and is for single-family households; the DeepSolar solar adopter count is for 2022, which is more recent and considers all rooftops. The relative percentage difference between LBNL's real solar adopter data~\cite{barbose2021residential} and our synthetic data is depicted in Figure~\ref{fig:validation_adopters}b. DeepSolar solar adopter data~\cite{wussow2024exploring} and our synthetic data are shown in Figure~\ref{fig:validation_adopters}c. As the LBNL dataset was also used as ground truth adopters, the discrepancy between the synthetic datasets and LBNL's real adopters is less compared to DeepSolar. In comparison between the LBNL dataset, 88\% of states have a difference under 25\%. Differences over 100\% in states like ND, NE, SD, and WV are capped at 100\% for visualization. These high discrepancies, especially in low-adoption areas, result in significant percentage differences. For instance, West Virginia's actual count of 62 versus a synthetic 338 leads to a 449\% difference. In comparison with the DeepSolar dataset, 60\% of the states have a difference under 50\%. Differences over 100\% in the DeepSolar dataset are in states IA, ID, ND, and NC. 

\begin{table}[!h]
\caption{Synthetic solar adopter comparison calibrated using DeepSolar and LBNL} 
\label{tab:calibration_deepsolar}
\begin{tabular*}{\textwidth}{@{\extracolsep\fill}lcccc}
\toprule
{States } & {DeepSolar} & {Synthetic\_DeepSolar} & {LBNL} & {Synthetic\_LBNL}\\
\midrule
AL&1490&1490&117&132\\
GA&5771&6109&1131&1151\\
IN&5638&5879&2783&2917\\
KS&2876&2974&1341&1407\\
MS&1199&1207&367&553\\
ND&277&279&21&1799\\
NE&973&935&424&1171\\
SD&336&360&25&555\\
TN&4208&4249&1490&1540\\
WV&1424&1607&62&338\\
\bottomrule
\end{tabular*}
\end{table}

Furthermore, we calibrated the ground truth adopter dataset from LBNL to DeepSolar and reran the solar adoption model in the ten states where the LBNL and DeepSolar vary significantly. This experiment demonstrates the framework's adaptability to calibrate the new ground truth dataset. Moreover, the models can be applied to any geographic region as they are generalizable. The results in Table~\ref{tab:calibration_deepsolar} further demonstrate this capability.

\subsection{: Algorithm and running time of our model}
\label{sec:apendixD}

Here, we describe our algorithms and runtime complexity for solar adoption and energy profile generation. The algorithm for the first step, i.e., predicting solar adopters, is presented in Algorithm~\ref{algo:task2a}. The runtime complexity of this process is $\mathcal{O}(m+h+p\log p)$, where $m$ denotes the model training and testing, $h$ denotes the hyperparameter tuning for the Gaussian process, and $p \log p$ is for sorting $X_{test}$ by expected improvement, considering $p$ points.  

\begin{algorithm}
\caption{ Solar adoption in synthetic population}
\label{algo:task2a}
\begin{flushleft}
\textbf{Input:} State $s$, Real world solar adopters count at state-level $s_{real,s}$, RECS file $f_{RECS}$, Synthetic population file for the state $f_{syn,s}$.

\textbf{Output:} A file $f_{syn,s,SOLAR}$, listing $H$ households, where each household $h_i$ is classified as either a solar adopter or not.
  
\end{flushleft}
\begin{algorithmic}[1]
\Procedure{MainFunction}{$s$,$s_{real,s}$,$f_{RECS}$,$f_{syn,s}$}
\Statex \textcolor{gray}{\textit{Feature selection and data preparation before running the ML model.}}
\State Read and prepare $f_{RECS}$ and $f_{syn,s}$. 
\Statex \textcolor{gray}{\textit{Selection of initial points for Bayesian optimization.}}
\State Select random indices from all combinations of $\beta (0.0-2.01)$and $\tau (0.05-0.95)$.
\For{each index in random indices}
    \State Perform model training, testing, and prediction with custom log loss function and custom $\tau$.
    \State Calculate $diff \gets |s_{real,s}-s_{syn,s}|$.
    %, where  $s_{syn,s}$ is the total count of synthetic solar adopter for given $s$.
\EndFor
\Statex \textcolor{gray}{\textit{Selection of point based on exploration and exploitation to get the  parameters that gives minimum discrepancy between the synthetic adopters and ground truth adopters.}}
\State Select $\beta$ and $\tau$ using Bayesian optimization and EI as the acquisition function, which gives $diff_{min}$.
\State Stop either at $\lvert diff \rvert \leq 0.15 \times s_{real,s}$ or for $2000$ rounds. 
\EndProcedure
\end{algorithmic}
\end{algorithm}

The algorithms for the second step are given in two steps: ($i$) Algorithm~\ref{algo:task3a} for time-invariant variables sample generation, and ($ii$) Algorithm~\ref{algo:task3b}. The first Algorithm is computed only once as these variables do not change w.r.t to time. The runtime complexity of Algorithm~\ref{algo:task3a} is $\mathcal{O}(N \cdot (n+P))$, where $N$ is the number of houses, $n$ is the number of samples generated and $P$ is a set of planes.

For Algorithm~\ref{algo:task3b}, the period $T$ depends on the type of input specified by the user. It can be a date $t$, year-month, or year-week combination. The runtime complexity of the framework is $\mathcal{O}(N + (((N \cdot W)/worker\_size) + \epsilon)\cdot T) $ where $N$ is the number of houses, $W$ is the total number of hours (24 hours),  $worker\_size$ is the size of the worker processes, $\epsilon$ denotes the communication overhead between the worker processes and main process and $T$ is a constant number based on the type of period input specified by the user. The size of worker processes depends on the number of nodes allocated based on the number of solar-adopted houses in $c$.

\begin{algorithm}
\caption{Time invariant variables sample generation}
\label{algo:task3a}
\begin{flushleft}
\textbf{Input:} State $s$, county $c$, a set of households $H$, a set of the location of the households denoted using its latitude, longitude ($lat_i$,$lon_i$) of household $h_i$, its house area $HA_i$ and its respective census tract $ct_i$.

\textbf{Output:} A file containing $N$ solar adopted households and for each $h_i$, its roof area $RA_i$, $type$, ($lat_i$,$lon_i$), $n$ samples for solar efficiency $\eta_i$, performance ratio $PR_i$, number of planes $P$, tilt $\theta_i$ and azimuth pairs $\omega_i$ and total area of the panels  $A_i$.
\end{flushleft}
\begin{algorithmic}[1]
\Procedure{MainFunction}{$s$,$c$,$ct$,$N$,$HA$,$lat$,$lon$}
\For{each $h_i$ in $N$}
    \Statex \textcolor{gray}{\textit{Compute roof area}}
    \State $RA_i \gets 1.5 \times HA$.
    \Statex \textcolor{gray}{\textit{Assign type of building based on roof area.}}
    \State $type \gets \text{small}$ \textbf{if} $RA_i \leq 464.6$ \textbf{else} $type \gets \text{medium}$.
    \Statex \textcolor{gray}{\textit{Get a set of values for solar efficiency and performance ratio following a uniform distribution.}}
    \State $\eta_i \sim \text{Uniform}(0.18, 0.22),PR_i \sim \text{Uniform}(0.5, 0.9)$.
    \Statex \textcolor{gray}{\textit{Get product of the two vectors.}}
    \State $rPR_i \gets np.outer(\eta_i, PR_i)$.
    \Statex \textcolor{gray}{\textit{Select set of number of planes based on weighted sampling according to the building type and weights present in the literature.}}
    \State $P \gets \text{Weighted sample}(\text{type}, \text{size}=n, \text{ weights=\text{~\cite{gagnon2016rooftop}}})$. 
    \For{each plane $p$ in $P$}
    \Statex \textcolor{gray}{\textit{Select appropriate fitted rate of decay of the exponential process (r) and optimal location parameter (t) values.}}
        \State $(r, t) \gets \begin{cases} (0.042, 10) & \text{if small and } p = 1, \\ (0.071, 10) & \text{if small and } p \neq 1, \\ (0.002, 300) & \text{if medium and } p = 1, \\ (0.046, 10) & \text{if medium and } p \neq 1. \end{cases}$
        \Statex \textcolor{gray}{\textit{Select a set of random values between 0 and the roof area of the building and  for each value calculate its possible weight.}}
        \State $V \sim \text{Uniform}(0, \text{roof area}),\, \text{scale} = \frac{1}{r},\, R \sim \text{Exp}(V,\text{scale}, \text{loc}=t)$
        \Statex \textcolor{gray}{\textit{Generate samples of available area based on the set of roof area and its corresponding weight.}}
        \State $S_i \sim \text{Sample}(\text{weights}=R, \text{size}=n)$
        \Statex \textcolor{gray}{\textit{Calculate the suitable area based on the available area and panel area.}}
        \State $A_i \gets \lfloor S_i/1.64 \rfloor \times 1.64$
    \EndFor
    \Statex \textcolor{gray}{\textit{Calculate the suitable area based on the available area and panel area.}}
    \Statex \textcolor{gray}{\textit{Select set of tilt and orientation pairs based on weighted sampling according to the building type and weights present in the literature.}}
    \State $(\theta_i, \omega_i) \sim \text{WeightedSample}(\text{type}, \text{size}=n, \text{weights}=\text{~\cite{gagnon2016rooftop}})$
\EndFor
\EndProcedure 
\end{algorithmic}
\end{algorithm}

\begin{algorithm}
\caption {Solar energy profile generation}
\label{algo:task3b}
\begin{flushleft}
\textbf{Input:} Time invariant variable sample generation file $f_1$, state $s$, county $c$, time period for generation $T$, a set of irradiance files $f_{ct}$ for all $ct$ in $c$.   

\textbf{Output:} File containing 24 hour profiles of $E_{\mu,i,d,w}$ for a given day $d$ and given hour $w$, its standard deviations $E_{\sigma,i,d,w}$, total daily $E_{\mu,i,d}$ and its standard deviation $E_{\sigma,i,d}$ for each household $h_i$
\end{flushleft}

\begin{algorithmic}[tbh]
\Procedure{MainFunction}{$f_1$,$s$,$c$,$f_{ct}$}
\Statex \textcolor{gray}{\textit{Process the time invariant pre-computed file.}}
\State Process $f_1$ file.
\Statex \textcolor{gray}{\textit{Get product of the suitable area and vector of values generated based on product of solar efficiency and performance ratio.}}
\State $ArPR_i \gets np.outer(A_i,rPR_i)$.
\Statex \textcolor{gray}{\textit{Based on temporal resolution value, calculate the dates for which the simulation needs to be performed.}}
\State Based on $T$, get the dates to be processed.
\For{each date in dates}
    \State \Call{ProcessDate}{$t$,$s$,$c$}
\EndFor   
\EndProcedure
\Procedure{ProcessDate}{$t$,$s$,$c$}
\State Extract the day, month and year from $t$ and calculate day of the year $d$.
\Statex \textcolor{gray}{\textit{Initialize MPI process and allocate houses to worker process based on the total number of houses.}}
\State Initialize MPI environment and allocate houses based on $worker\_size$.
\Statex \textcolor{gray}{\textit{Call the function to compute the hourly solar values.}}
\For {each $h_i$ in range($start\_index$, $end\_index$)}
    \State \Call{ComputeHourlySolarEnergy}{$f_{ct}$, $\theta_i$, $\omega_i$, $lat_i$,$d$,$t$, $ArPR_i$} 
\EndFor
\Statex \textcolor{gray}{\textit{Master process aggregates the results from the worker process.}}
\State Master process aggregates and finalizes output.
\EndProcedure
\Procedure{ComputeHourlySolarEnergy}{$f_{ct}$, $\theta_i$, $\omega_i$, $lat_i$,$d$,$t$, $ArPR_i$}
\Statex \textcolor{gray}{\textit{Get the global horizontal irrandiance information based on census tract.}}
\State Get the $GHI$ hourly values for the specific day from $f_{ct,i}$ of the household.
\For {each hour $w$ in $W$}
\Statex \textcolor{gray}{\textit{Compute the hourly solar radiation on a  tilted panels.}}
    \State Calculate all $H_{i,d,w},$ values depending on $GHI_{i,d,w}$, $\theta_i$ and $\omega_i$~\cite{duffie2020solar}.
    \Statex \textcolor{gray}{\textit{Compute the product of two vectors to calculate hourly energy produced.}}
    \State $E_{i,d,w} \gets np.outer(H_{i,d,w},ArPR_i)$.
    \Statex \textcolor{gray}{\textit{Compute the mean energy produced and its standard deviation.}}
    \State $E_{\mu,i,d,w} \gets \mu (E_{i,d,w})$, $E_{\sigma,i,d,w} \gets \sigma (E_{i,d,w})$.
\EndFor
\Statex \textcolor{gray}{\textit{Aggregate the results for a day and compute its mean and standard deviation.}}
\State $E_{\mu,i,d} \gets \sum E_{\mu,i,d,w}$, $E_{\sigma,i,d} \gets \sqrt{\sum (E_{\sigma,i,d,w}^2)}$.
\State \Return $h_i$, $E_{\mu,i,d,w}$, $E_{\sigma,i,d,w}$, $E_{\mu,i,d}$, $E_{\sigma,i,d}$.
\EndProcedure 
\end{algorithmic}
\end{algorithm}